\documentclass{iopjournal}
\usepackage{amsmath}
\usepackage{amssymb}
\usepackage[section]{placeins}

\makeatletter

\renewcommand{\articletype}[1]{\vspace*{-2mm}\noindent{\scriptsize \sf{\bfseries \MakeUppercase{#1}}}\vspace{2mm}}

\renewcommand{\title}[1]{{\exhyphenpenalty=10000\hyphenpenalty=10000 
 \fontsize{18}{21}\selectfont\noindent\raggedright
        \textsf{#1}\par}}
\makeatother

\renewcommand{\headrulewidth}{0pt}

\graphicspath{{Images/}}

\begin{document}

\articletype{Paper}

\title{How Much Capacity Does EEG Denoising Need? Ultra-Compact Networks reveal Benchmark Saturation and Metric-Utility Gap}

\author{Jasmeet Singh Bindra$^{1,*}$\orcid{0009-0006-3105-0758}, Siddharth Panwar$^2$\orcid{0000-0002-9432-350X}}

\affil{$^1$Indian Knowledge Systems and Mental Health Applications (IKSMHA) Center, Indian Institute of Technology Mandi, Himachal Pradesh, India}

\affil{$^2$School of Computing and Electrical Engineering, Indian Institute of Technology Mandi, Himachal Pradesh, India}

\affil{$^*$Author to whom any correspondence should be addressed.}

\email{D24059@students.iitmandi.ac.in; siddharthpanwar@iitmandi.ac.in}


\begin{abstract}
\textbf{Objective.} Deep Learning based EEG denoising architectures have scaled from tens of thousands to tens of millions of parameters, yet to our knowledge, no prior study has isolated model capacity as the experimental variable or tested whether the reconstruction metrics these models optimize predict downstream neural-signal utility. We address both gaps.
\textbf{Approach.} We kept architecture, loss, data split, and training recipe fixed and swept only channel width from 1.05K to 40.26K trainable parameters in a minimal depthwise-separable convolutional U-Net. Models were evaluated on the EEGDenoiseNet benchmark, a million-segment mixed-artifact corpus, two cross-dataset BCI zero-shot transfer tests, controlled retraining of established baselines under the same pipeline, and downstream motor-imagery classification with five decoder families across all nine BCI Competition IV-2a subjects.
\textbf{Main Results.} Reconstruction performance saturated by 3-6.5K parameters across all benchmark settings, with post-elbow gains of at most 0.015 correlation coefficient (CC) per log$_{10}$-parameter unit. An 8.46M-parameter EEGDN CNN retrained under the same pipeline matched the 40.26K parameter compact variant on EOG, a 200$\times$ parameter gap yielding no reconstruction advantage, while a second-family Patch-Transformer control reproduced the same diminishing-return shape. Downstream evaluation exposed a classifier-dependent metric--utility gap: reconstruction-optimized denoising significantly degraded CSP+LDA motor-imagery classification across all nine subjects and three artifact types (best denoised accuracy 0.547 versus 0.612 noisy baseline; Bonferroni $p=0.0488$), with the gap persisting on naturally recorded trials ($\Delta=-0.047$; BH-FDR $q=0.0049$). End-to-end neural decoders showed variable or neutral effects, indicating that the gap is classifier-dependent.
\textbf{Significance.} Standard EEG denoising benchmarks are saturated far below the capacity of current models, and the reconstruction metrics they reward do not predict downstream BCI utility. Ultra-compact models at 33-46 KB and 1.27-2.61M FLOPs per segment are practical for edge EEG deployment. These findings argue for capacity-controlled evaluation, harder benchmarks with task-aware endpoints, and mandatory downstream validation in future denoising studies.
\end{abstract}

\section{Introduction}

EEG denoising architectures have grown from benchmark CNN/RNN baselines to U-Nets, adversarial hybrids, Transformers, and state-space models, with parameter counts spanning tens of thousands to tens of millions \cite{zhang2021eegdenoisenet,chuang2022icunet,dong2023wgan,tang2025ctdcenet,chen2025denoisemamba}. This growth has improved reconstruction metrics on semi-synthetic benchmarks, but two basic questions remain unresolved: how much capacity do these benchmarks actually require, and do reconstruction metrics predict downstream EEG utility?

EEG is central to brain--computer interfaces (BCIs), clinical neuroscience, cognitive monitoring, and neurotechnology because it is non-invasive, portable, and temporally precise. These same recordings are vulnerable to non-neural contamination from ocular, muscular, cardiac, line-noise, and electrode-related sources \cite{croft2000ocular,urigueen2015artifact}. Classical methods such as filtering, regression, ICA/BSS, artifact subspace reconstruction, and automated rejection remain valuable \cite{gratton1983ocular,makeig1996ica,jung2000bss,mullen2015realtime,chang2020asr,nolan2010faster,mognon2011adjust}, but heterogeneous artifacts and spectrally overlapping neural activity have motivated learned denoising models.

This raises two connected questions. First, how much model capacity does standard EEG denoising actually require? Existing comparisons typically evaluate one proposed architecture against a set of baselines, but they do not isolate capacity as the experimental variable. As a result, it is difficult to determine whether improved performance comes from a specific architectural mechanism, a larger parameter budget, training details, or benchmark saturation. Second, do reconstruction metrics measure the form of signal preservation that matters for downstream EEG use? Correlation coefficient (CC), root mean square error (RMSE), signal-to-distortion ratio (SDR), and related temporal or spectral errors quantify similarity to an assumed-clean reference. These metrics are essential for standardized benchmarking, but they do not directly test whether denoising preserves task-relevant neural features. This distinction is especially important because supervised EEG denoising benchmarks generally treat preprocessed EEG as the clean target. A model optimized to reproduce that target may achieve high reconstruction fidelity while still suppressing neural components that are useful for classification, detection, or interpretation.

We hold the architecture fixed and sweep only channel width from 1.05K to 40.26K parameters, evaluating reconstruction across EEGDenoiseNet, a large mixed-artifact corpus, BCI zero-shot transfer, controlled baseline retraining, compute profiling, and five downstream BCI decoder families. This design turns model capacity into the experimental variable rather than a hidden property of unrelated architectures.

Our contributions are threefold:
\begin{enumerate}
    \item A controlled width sweep shows that reconstruction performance saturates by 3--6.5K parameters on the standard EEGDenoiseNet benchmark, a compositionally harder million-segment mixed-artifact corpus, and two cross-dataset BCI zero-shot transfer tests, with the pattern supported by a second-family Patch-Transformer control and same-pipeline EEGDN CNN and MicroWaveNet retraining.
    \item Reconstruction-optimized denoising significantly degrades Common Spatial Patterns (CSP) + Linear-Discriminant Analysis (LDA) classification across all nine BCI IV-2a subjects and three artifact types (\(p=0.0098\), Bonferroni \(p=0.0488\) for synthetic EOG), with the gap persisting on real non-synthetic recordings, while end-to-end neural decoders show variable effects.
    \item Base4--base6 models occupy 33--46 KB and require 1.27--2.61M FLOPs per segment, identifying deployment-ready operating points for embedded EEG systems.
\end{enumerate}

Together, these results argue for capacity-controlled denoising evaluation, harder benchmarks, and task-aware validation that tests whether denoising preserves the neural information downstream applications require.

The capacity ceiling and metric--utility gap have direct implications for neural signal processing pipelines: practitioners risk degrading BCI performance by adopting reconstruction-optimized denoisers without downstream validation.

\section{Related Work}

\subsection{Architectural Progression Without Capacity Control}

Deep EEG denoising has progressed through increasingly specialized architectures, paralleling the broader growth of deep EEG analysis \cite{roy2019systematic}. EEGDenoiseNet established a standard supervised setting in which a contaminated single-channel segment is mapped to an assumed-clean target and evaluated with reconstruction metrics \cite{zhang2021eegdenoisenet}. Its CNN and RNN baselines (hereafter EEGDN CNN and EEGDN RNN) remain important reference points because they place early benchmark models at 8.40M and 787.0K parameters in our comparison, far above the ultra-compact regime.

Subsequent work introduced encoder-decoder, U-Net, generative, recurrent, Transformer, and state-space mechanisms. IC-U-Net and LRR-UNet extended U-Net-style recovery \cite{chuang2022icunet,yue2025lrrunet}; AR-WGAN, DuoCL, and GCTNet introduced adversarial, recurrent, and hybrid CNN-Transformer components \cite{dong2023wgan,gao2023duocl,yin2025gctnet}; and Denosieformer, CT-DCENet, and DenoiseMamba represent the high-capacity end of the progression \cite{chen2024denosieformer,tang2025ctdcenet,chen2025denoisemamba}. Parameter counts at this end range from hundreds of thousands to tens of millions (e.g., EEGDN CNN 8.40M, CT-DCENet 38.10M), but most of these models lack publicly available code, making independent reproduction of their reported metrics difficult.

Lightweight models provide a counterpoint but still do not isolate capacity. DeepSeparator and MicroWaveNet occupy the 32--60K range and show that denoising does not always require million-parameter models \cite{yu2022deepseparator,lahiri2025microwavenet}. LTDNet-EEG similarly targets portable and wearable deployment \cite{huang2024ltdnet}. These studies compare proposed architectures against baselines, but they do not ask whether gains arise from architectural mechanisms, larger parameter budgets, training recipes, or benchmark saturation. Despite the architectural progression, no systematic study has examined the minimum capacity required for standard EEG denoising benchmarks or whether reconstruction metrics predict downstream task utility.

\subsection{Efficiency, Scaling, and Capacity Analysis}

Efficiency-aware modeling is well established in EEG classification and broader neural-network design. EEGNet showed that depthwise and separable convolutions can produce compact EEG classifiers across BCI paradigms \cite{lawhern2018eegnet}, while MobileNets, Xception, ECA-Net, and deep compression established efficient convolution, attention, pruning, and quantization patterns for low-compute models \cite{chollet2017xception,howard2017mobilenets,wang2020ecanet,han2016deepcompression}. Classical online cleaning and preprocessing pipelines such as artifact subspace reconstruction and PREP also reflect deployment-oriented EEG practice \cite{mullen2015realtime,chang2020asr,bigdely2015prep}. In denoising, compressed and lightweight approaches exist \cite{nagar2021compressed,yu2022deepseparator,lahiri2025microwavenet}, but most studies still emphasize new modules or reconstruction gains rather than controlled capacity.

The capacity question connects directly to broader scaling and over-parameterization theory. Neural scaling laws show that model size, data, and compute can follow predictable performance trends \cite{kaplan2020scaling}, while the lottery ticket hypothesis shows that sparse subnetworks can match dense networks when trained appropriately \cite{frankle2019lottery}. These literatures treat capacity as an experimental variable. EEG denoising has not: the field has lacked a capacity-controlled sweep that asks how little model is sufficient for the benchmark.

\subsection{Semi-Synthetic Benchmarks and Their Assumptions}

EEGDenoiseNet is the de facto benchmark for supervised single-channel EEG denoising \cite{zhang2021eegdenoisenet}. It contains 4514 clean EEG segments, 3400 ocular artifact segments, and 5598 muscular artifact segments sampled at 512 Hz. Noisy examples are generated semi-synthetically by linearly mixing clean EEG with artifact segments at controlled signal-to-noise ratios, allowing each contaminated input to be paired with a clean reference target. This construction makes it possible to report standardized reconstruction metrics such as correlation coefficient and relative root mean square error across EOG and EMG artifact conditions.

Its controlled design has made EEGDenoiseNet widely adopted by subsequent work, including DeepSeparator \cite{yu2022deepseparator}, Denosieformer \cite{chen2024denosieformer}, CT-DCENet \cite{tang2025ctdcenet}, DenoiseMamba \cite{chen2025denoisemamba}, and MicroWaveNet \cite{lahiri2025microwavenet}. The same design encodes a key assumption: segments labeled as ``clean'' are treated as ground truth, although in practice clean EEG reflects quality-control and preprocessing decisions rather than direct access to artifact-free neural sources \cite{delorme2004eeglab,nolan2010faster,mognon2011adjust}. This assumption is appropriate for standardized reconstruction benchmarking, but it limits what reconstruction metrics alone can establish about functional signal utility.

Real-artifact corpora expose the complementary problem. The Temple University Hospital EEG corpus provides clinical recordings with real, heterogeneous artifact conditions \cite{obeid2016temple}, but such recordings generally lack paired clean-reference targets, making them incompatible with standard supervised reconstruction evaluation. This gap between the benchmarks the field uses and the conditions it must ultimately address motivates downstream task-aware validation alongside clean-reference metrics.

\section{Methods}

\subsection{Architecture}

We model single-channel EEG denoising as a supervised reconstruction problem. Throughout this paper, lowercase letters denote the model-level signals where: \(x \in \mathbb{R}^{T}\) is the assumed-clean EEG segment,
\(a \in \mathbb{R}^{T}\) the artifact contribution, and \(y = x + a\) the
contaminated input. Uppercase letters \(Y\), \(X\), and \(A\) denote the
corresponding stored arrays. All signals are
normalized by the standard deviation of the contaminated signal,
\(\sigma_y = \mathrm{std}(y)\). The denoiser \(f_{\theta,C}\) maps \(y\) to a
clean estimate \(\hat{x} = f_{\theta,C}(y)\), where \(C\) is the base channel
width.

\begin{figure}[tbp]
\centering
\includegraphics[width=\textwidth]{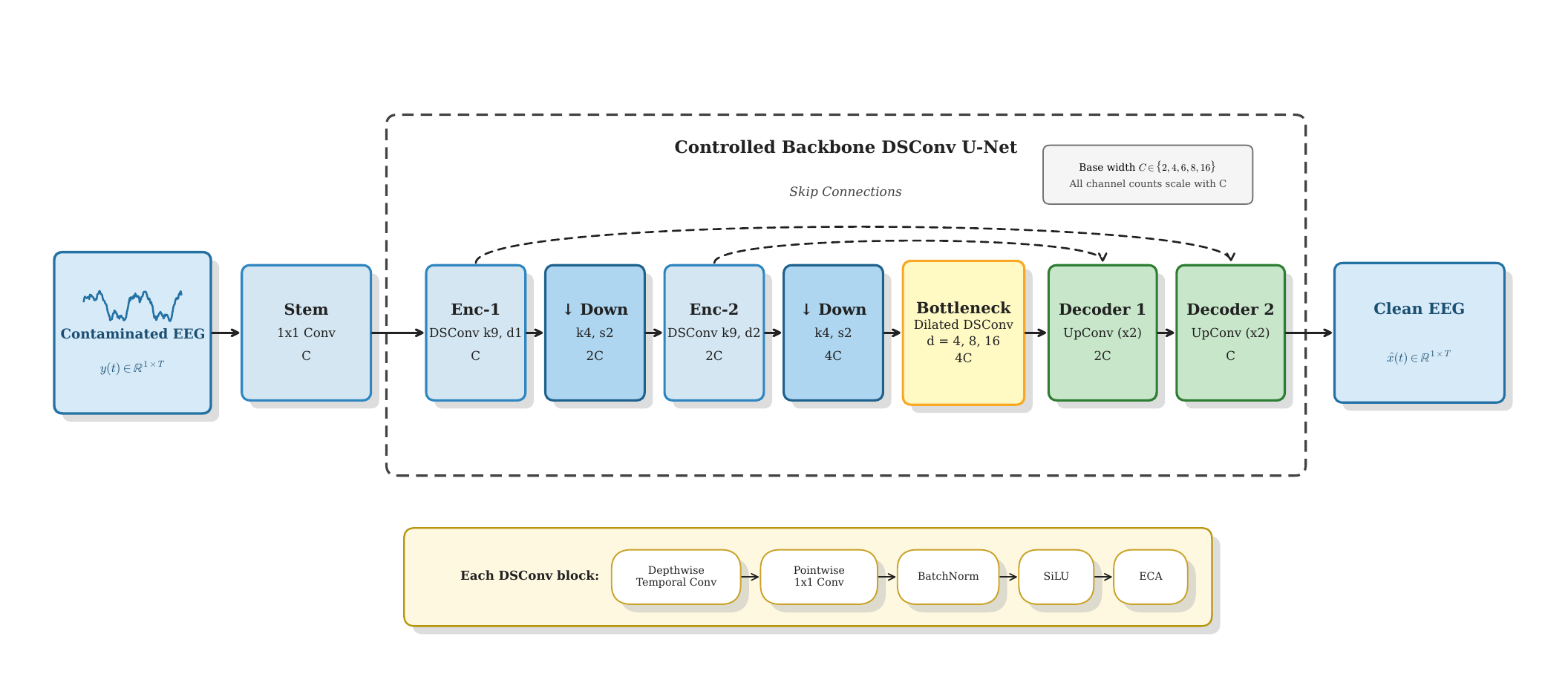}
\caption{Architecture of the clean-only depthwise-separable convolutional denoiser. The base width \(C\) sets the channel count in the stem, encoder, bottleneck, and decoder. The final width-sweep model uses two downsampling stages, a dilated bottleneck, symmetric decoder stages, skip connections, and ECA attention within depthwise-separable convolution blocks.}
\label{fig:architecture}
\end{figure}

The architecture used for the width sweep is the controlled backbone
shown in Figure~\ref{fig:architecture}. The backbone is intentionally minimal: a standard depthwise-separable convolutional U-Net with efficient channel attention, designed to isolate channel capacity as the sole experimental variable rather than to propose a new architecture. To ensure that only channel capacity varied across experimental conditions, the backbone uses a single clean-output head without auxiliary branches or losses. It is a one-dimensional U-Net-like temporal encoder--decoder built from
depthwise-separable convolution blocks with efficient channel attention (ECA),
following efficient convolution and lightweight channel-attention design
principles \cite{chollet2017xception,wang2020ecanet}.
The model takes an input tensor of shape \(1 \times T\) and produces one output
channel, \(\hat{x}\).

The controlled backbone design was motivated by an additive component study (Supplementary Table S11, Supplementary Figure S6a). Among four variants at 40.24--40.26K parameters, only ECA attention provided a small improvement (\(\Delta=+0.005\) CC on EMG, \(n=3\) seeds), and it is retained as a low-cost channel-attention mechanism. Auxiliary artifact-head, spectral, and mixture losses were neutral or harmful and were excluded to maintain the single-variable controlled design.

The network begins with a \(1\times1\) stem convolution from 1 channel to \(C\)
channels. The encoder contains two depthwise-separable convolution blocks and
two strided downsampling convolutions. The first encoder block maps
\(C\rightarrow C\) with kernel size 9 and dilation 1. The first downsampling
layer is a bias-free convolution with kernel size 4, stride 2, and padding 1,
mapping \(C\rightarrow2C\). The second encoder block maps \(2C\rightarrow2C\)
with kernel size 9 and dilation 2. The second downsampling layer has the same
kernel, stride, and padding and maps \(2C\rightarrow4C\). The bottleneck contains
three depthwise-separable convolution blocks at width \(4C\), with kernel size 9
and dilations 4, 8, and 16. The decoder upsamples with two bias-free transposed
convolutions, each using kernel size 4, stride 2, and padding 1. The first maps
\(4C\rightarrow2C\), is concatenated with the corresponding \(2C\) encoder skip
feature, and is followed by a \(4C\rightarrow2C\) depthwise-separable block with
dilation 2. The second maps \(2C\rightarrow C\), is concatenated with the
\(C\)-channel encoder skip, and is followed by a \(2C\rightarrow C\) block with
dilation 1. A final \(1\times1\) convolution maps \(C\rightarrow1\). All
depthwise-separable blocks use batch normalization, Sigmoid Linear Unit (SiLU) activation, ECA
attention, and an identity residual connection only when the input and output
channel counts match.

For an input feature map \(H\in\mathbb{R}^{C_{\mathrm{in}}\times T}\), the
depthwise-separable convolution first applies a depthwise temporal convolution
independently to each channel,
\begin{equation}
u_i(t)=\sum_{r=0}^{k-1}w^{\mathrm{dw}}_{i,r}
H_i\left(t+d(r-\lfloor k/2\rfloor)\right),
\end{equation}
where \(k=9\) and \(d\) is the block dilation. A pointwise convolution then mixes
channels,
\begin{equation}
v_j(t)=\sum_{i=1}^{C_{\mathrm{in}}}w^{\mathrm{pw}}_{j,i}u_i(t).
\end{equation}
The block output before residual addition is
\begin{equation}
z=\mathrm{ECA}\left(\mathrm{SiLU}\left(\mathrm{BN}(v)\right)\right).
\end{equation}

ECA computes a channel descriptor by global temporal averaging,
\begin{equation}
s_c=\frac{1}{T}\sum_{t=1}^{T} z_c(t),
\end{equation}
applies a one-dimensional convolution of kernel size 3 along the channel axis,
and gates the features with a sigmoid:
\begin{equation}
g=\sigma(\mathrm{Conv1D}_{k=3}(s)), \qquad
\tilde{z}_c(t)=g_c z_c(t).
\end{equation}

\subsection{Width Parameterization}

For the controlled-backbone, the base width \(C\) controls every channel dimension in the encoder, bottleneck,
and decoder. We evaluated \(C\in\{2,4,6,8,10,12,16\}\), producing models with
1.05K, 3.22K, 6.53K, 10.99K, 16.59K, 23.34K, and 40.26K trainable parameters,
respectively. Because the depth, kernel sizes, dilation schedule, attention
kernel, normalization, activation function, optimizer, and loss were fixed, this
family isolates the effect of channel capacity.

\subsection{Datasets}

\subsubsection{EEGDenoiseNet benchmark}

We used the EEGDenoiseNet semi-synthetic benchmark protocol
\cite{zhang2021eegdenoisenet}. For EOG contamination, the prepared dataset
contains 27,200 training, 3,400 validation, and 3,400 test segments. Each segment
has \(T=512\) samples at 256 Hz. The split is 80/10/10 before expansion, with
3,400 clean EEG and 3,400 EOG segments mixed across 10 SNR levels from
\(-7\) to \(2\) dB. For EMG contamination, the prepared dataset contains 53,736
training, 6,708 validation, and 6,732 test segments. Each segment has
\(T=1024\) samples at 512 Hz. The EMG preparation uses 4,514 EEG segments and
5,598 EMG segments, reusing EEG as needed to match the artifact pool, with 12
SNR levels from \(-7\) to \(4\) dB.

For all semi-synthetic datasets, the mixture coefficient was chosen from the
desired SNR \(r\) as
\begin{equation}
\lambda=\sqrt{\frac{P_x}{P_a 10^{r/10}}},
\qquad
y=x+\lambda a,
\end{equation}
where \(P_x\) and \(P_a\) are the mean squared powers of the clean and artifact
segments.

\subsubsection{Large-scale mixed-artifact corpus}

The mixed-artifact corpus contains 1,000,000 segments at 256 Hz, each with
\(T=512\) samples. It was generated from EEGDenoiseNet source pools:
4,514 clean EEG segments of 512 samples, 3,400 EOG artifact segments of
512 samples, and 5,598 EMG artifact segments of 1,024 samples downsampled from
512 Hz to 256 Hz. Source-level splitting was performed before any mixing:
clean EEG, EOG, and EMG source indices were assigned to disjoint train,
validation, and test pools, preventing source segments from appearing in more
than one split. The final dataset contains 800,000 training, 100,000 validation,
and 100,000 test mixtures, stored as 80, 10, and 10 chunks of 10,000 segments.
The mixed-1M corpus uses the same underlying EEG and artifact source pools as
EEGDenoiseNet but substantially extends the evaluation challenge: it introduces
multi-artifact mixtures (EOG+EMG, line noise, ECG, electrode drift) not present
in the original benchmark, a harder SNR distribution with 30\% of segments below
\(-7\) dB, stochastic amplitude scaling, and 10$\times$ more test segments for
tighter statistical estimation. Results on this corpus test robustness to
compositional artifact complexity and more adverse noise conditions within the
same source distribution.

SNRs were sampled from a two-part distribution: 70\% of mixtures used
\(\mathrm{SNR}\sim\mathcal{U}[-7,2]\) dB and 30\% used
\(\mathrm{SNR}\sim\mathcal{U}[-12,-7]\) dB. Seven artifact recipes were sampled
with fixed probabilities: EOG (0.25), EMG (0.25), EOG+EMG (0.20), EMG+LINE
(0.10), EOG+LINE (0.10), EOG+EMG+LINE (0.05), and EOG+EMG+LINE+ECG (0.05).
EOG segments were polarity-flipped with probability 0.5 and pre-scaled by
\(\mathcal{U}[0.7,1.3]\); EMG segments were pre-scaled by
\(\mathcal{U}[0.6,1.5]\). Electrode noise was added with probability 0.35.
Line noise used a 50 Hz sinusoid with probability 0.85 and a 60 Hz sinusoid
otherwise, with harmonics and slow amplitude modulation. ECG contamination was
generated as a synthetic QRS pulse train with an optional T-wave component.
After artifact components were combined, the final artifact mixture was scaled
to the target SNR.

\subsubsection{BCI zero-shot denoising datasets}

We used BCI Competition IV-2a and IV-2b recordings as zero-shot denoising tests
\cite{tangermann2012bci}. This protocol is not downstream classification. Real
BCI EEG provides the assumed-clean target, real BCI EOG provides the artifact
template, and known-SNR mixtures are generated so reconstruction metrics can be
computed.

For BCI IV-2a, we used subject A01T, EEG channel index 9, and EOG channel index
22. The recording was resampled from 250 Hz to 256 Hz. Clean windows were drawn
from the lowest 20\% of EOG-band energy and artifact windows from the highest
10\%, giving 2,148 clean starts and 1,074 artifact starts. For BCI IV-2b, we
used B0101T, EEG:C3, and EOG:ch01, giving 1,934 clean starts and 967 artifact
starts after the same selection rule. For each dataset, 5,000 test mixtures were
generated with seed 42 and SNR uniformly sampled from \([-6,2]\) dB.

\subsection{Training Protocol}

All width variants were trained with AdamW, initial learning rate
\(10^{-3}\), weight decay \(10^{-4}\), gradient clipping at norm 1.0, and
checkpoint selection by validation SDR. EEGDenoiseNet EOG and EMG width-sweep
runs used 25 epochs, batch size 256, two data-loading workers, and a 3-epoch
linear warm-up followed by cosine decay. Mixed-1M runs used 10 epochs, batch size
1024, two workers, and a 2-epoch linear warm-up followed by cosine decay. The
main multi-seed experiments used seeds 42, 43, and 44. For the final
EEGDenoiseNet Pareto-width statistical aggregation, base2, base4, base6, base8,
and base16 were additionally trained with seeds 45 and 46, giving five seeds
\(\{42,43,44,45,46\}\) for those widths. Base10 and base12 remained
exploratory \(n=3\) widths.

The training objective was clean reconstruction only:
\begin{equation}
\mathcal{L}_{x}
=
\frac{1}{T}\sum_{t=1}^{T}
\sqrt{(\hat{x}(t)-x(t))^2+\epsilon_H^2},
\qquad \epsilon_H=2\times10^{-3}.
\end{equation}
The pseudo-Huber form with \(\epsilon_H = 2 \times 10^{-3}\) transitions smoothly
from L2 behaviour for small residuals to L1 behaviour for large residuals,
providing smooth gradients near zero while maintaining robustness to outliers
\cite{charbonnier1994}.
Auxiliary artifact, mixture-consistency, spectral, and envelope losses were set
to zero for the width-sweep experiments.

All full training and profiling runs were executed on the AI Lab workstation
with an AMD EPYC 7313 16-core CPU, 32 logical threads, an NVIDIA RTX A5000 GPU
with 24,564 MiB memory, PyTorch 2.10.0+cu128, and CUDA 12.8.

\subsection{Evaluation Metrics}

Let \(\hat{x}\) be the denoised estimate and \(x\) the assumed-clean reference.
Correlation coefficient (CC) was computed after temporal mean removal:
\begin{equation}
\mathrm{CC}(x,\hat{x})
=
\frac{\sum_{t=1}^{T}(x_t-\bar{x})(\hat{x}_t-\bar{\hat{x}})}
{\sqrt{\sum_{t=1}^{T}(x_t-\bar{x})^2}
 \sqrt{\sum_{t=1}^{T}(\hat{x}_t-\bar{\hat{x}})^2}}.
\end{equation}
Root mean squared error was
\begin{equation}
\mathrm{RMSE}(x,\hat{x})
=
\sqrt{\frac{1}{T}\sum_{t=1}^{T}(x_t-\hat{x}_t)^2}.
\end{equation}
Temporal relative RMSE was
\begin{equation}
\mathrm{T\mbox{-}RRMSE}(x,\hat{x})
=
\frac{\lVert x-\hat{x}\rVert_2}{\lVert x\rVert_2}.
\end{equation}
For spectral metrics, Welch power spectral densities \(P_x(f)\) and
\(P_{\hat{x}}(f)\) were estimated with \(n_{\mathrm{perseg}}=\min(256,T)\).
Spectral relative RMSE was
\begin{equation}
\mathrm{S\mbox{-}RRMSE}(x,\hat{x})
=
\frac{\lVert P_x-P_{\hat{x}}\rVert_2}{\lVert P_x\rVert_2}.
\end{equation}
Signal-to-distortion ratio was
\begin{equation}
\mathrm{SDR}(x,\hat{x})
=
10\log_{10}
\frac{\sum_{t=1}^{T}x_t^2+\epsilon}
{\sum_{t=1}^{T}(x_t-\hat{x}_t)^2+\epsilon}.
\end{equation}
In both SDR and PSD-KLD, \(\epsilon=10^{-10}\) serves as a numerical stabilization constant. For PSD-KLD, PSDs were normalized to probability vectors
\(p_x(f)\) and \(p_{\hat{x}}(f)\), and
\begin{equation}
\mathrm{PSD\mbox{-}KLD}(x,\hat{x})
=
\sum_f p_x(f)
\log\frac{p_x(f)+\epsilon}{p_{\hat{x}}(f)+\epsilon}.
\end{equation}

\subsection{Downstream BCI Classification Protocol}

Downstream utility was evaluated on BCI Competition IV-2a using all nine
subjects (A01--A09). For each subject, the official A0xT session was used for
training and the A0xE session was used as the held-out evaluation session. No
cross-validation was used; the protocol followed the fixed train/test split.
The first 22 EEG channels were used for classification, and EOG channel index 22
was used only to generate synthetic contamination. Trials labeled as artifacts
were excluded. Trials were cropped from 2.5 s to 6.0 s after cue onset and
resampled from 250 Hz to 256 Hz, giving 896 samples per trial. EOG-contaminated
versions of the training and test trials were generated with SNR uniformly
sampled from \([-6,2]\) dB. The test contamination used seed 42 and the training
contamination used seed 1042.
For contamination-type robustness, we repeated the CSP+LDA matched
denoised/denoised protocol with EOG, EMG, and EOG+EMG+LINE recipes. EOG
templates came from the BCI IV-2a EOG channel, EMG templates came from the
EEGDenoiseNet raw EMG artifact pool after resampling to 256 Hz, and mixed
artifacts combined EOG, EMG, and synthetic 50/60 Hz line noise before global
SNR scaling.

We evaluated five decoder families. The classical spatial-filter pipeline used
CSP+LDA, a standard linear approach for motor-imagery BCI
\cite{pfurtscheller2001motor,ramoser2000csp,blankertz2008spatial,lotte2018review}.
This pipeline applied a fourth-order Butterworth bandpass filter from 8 to
30 Hz. CSP used 8 components, Oracle Approximating Shrinkage (OAS) covariance regularization, log-variance
features, per-class covariance estimated from time-concatenated trials
(rather than averaging per-trial covariance matrices), and no trace
normalization. The classifier was LDA with LSQR solver and automatic shrinkage. Because CSP and LDA are linear feature-extraction and classification stages, changes in classification accuracy were interpreted as task-level utility changes rather than as direct localization of neural generators \cite{parra2005recipes,haufe2014interpretation}.

The remaining four decoders: EEGNet \cite{lawhern2018eegnet}, ShallowFBCSPNet, Deep4Net \cite{schirrmeister2017deep}, and EEGConformer \cite{song2023eegconformer}, used official Braindecode implementations \cite{schirrmeister2017deep} and were trained for 80 epochs with AdamW, learning rate \(10^{-3}\), weight decay \(10^{-4}\), and batch size 64. A 200-epoch convergence audit is reported in the supplementary material.

For each classifier, we evaluated five train/test conditions: clean/clean,
clean/noisy, noisy/noisy, clean/denoised, and denoised/denoised. The
clean/denoised condition tests strict preservation under a clean-trained
classifier, whereas denoised/denoised tests the matched practical case in which
the classifier is trained and tested on the denoiser output. Each of the 22 EEG channels was denoised independently by the single-channel model, matching the single-channel benchmark protocol. To test whether this per-channel application introduces cross-channel covariance effects, we additionally trained multichannel controls (Section~\ref{sec:downstream}). Statistical
significance of the matched denoised/denoised versus noisy/noisy comparison was
assessed with one-sided Wilcoxon signed-rank tests across subjects, with the
alternative hypothesis that denoised accuracy is lower than noisy/noisy
accuracy.
To test this single-channel-to-multichannel confound directly, we additionally
trained subject-specific 22-channel Depthwise-Separable Convolution (DSConv) U-Net controls at base6 and base16 on the BCI IV-2a training sessions and evaluated matched CSP+LDA on the held-out
evaluation sessions. These multichannel controls were used only as downstream
diagnostics and were not part of the main single-channel reconstruction
benchmark.

\subsection{Compute Profiling Methodology}

Compute profiling was performed on 512-sample inputs at 256 Hz. FLOPs were computed as \(2\times\)MACs counted with hooks on Conv1d, ConvTranspose1d, Linear, and LSTM layers. Latency was measured on CPU and CUDA for batch sizes 1 and 256 using 30 warm-up and 100 timed iterations with device synchronization. The full profiling methodology is provided in the supplementary material.

\section{Results}

Unless stated otherwise, results for our models are reported as mean \(\pm\) sample standard deviation over independent training seeds. Mixed-1M and BCI zero-shot evaluations use seeds 42, 43, and 44; the final EEGDenoiseNet Pareto widths use seeds 42--46. Published external baselines are reported as single reference values.

\subsection{Reconstruction Performance Saturates Early}
\label{sec:reconstruction_saturation}

The lower-anchor baselines show that EEGDenoiseNet requires learning, but not large capacity. Zero-parameter filters reached at most EOG CC 0.552 and EMG CC 0.743, a 34-parameter learned FIR reached 0.784/0.779, and a 365-parameter tiny TCN reached 0.849/0.797 (Supplementary Table S3). The controlled width sweep then pushed reconstruction into the benchmark-saturating regime with only a few thousand parameters.

\begin{figure}[tbp]
\centering
\includegraphics[width=\textwidth]{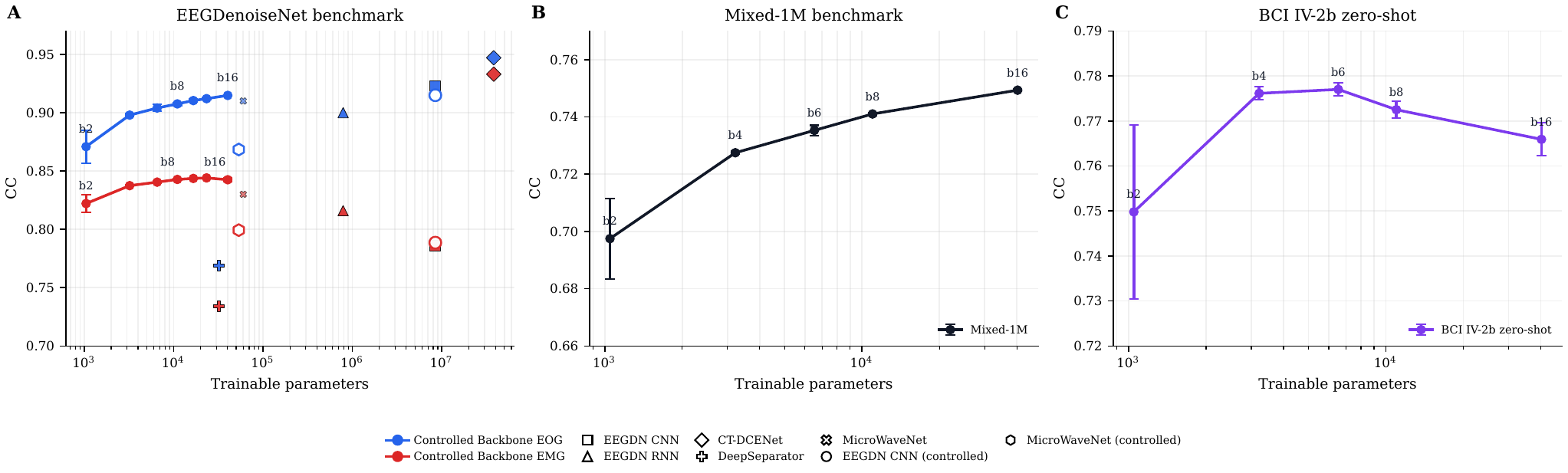}
\caption{Pareto frontier across reconstruction benchmarks. Panel A shows EEGDenoiseNet EOG and EMG CC versus parameter count, Panel B shows Mixed-1M CC, and Panel C shows BCI IV-2b zero-shot CC. Controlled backbone variants are shown as connected width-sweep points, and external baselines are shown as reference markers. Parameter count is shown on a log scale.}
\label{fig:pareto_frontier}
\end{figure}

On EEGDenoiseNet, most reconstruction gain appeared below 10K parameters (Table~\ref{tab:eegdenoisenet}; Figure~\ref{fig:pareto_frontier}A). Base8 reached EOG CC \(0.907 \pm 0.001\) with 10.99K parameters, while base16 reached \(0.915 \pm 0.001\) with 40.26K parameters. EMG saturated earlier: base6--base16 differed by only 0.003 CC. A segmented regression of CC against \(\log_{10}\) parameter count selected base4 as the descriptive elbow for both EOG and EMG. The post-elbow slope was \(0.0151\) CC per log10-parameter unit for EOG and \(0.0044\) for EMG, with the EMG bootstrap interval including zero (\(-0.0003\) to \(0.0056\)). Supplementary Table S1 gives the fixed \(n=5\) statistics, Supplementary Table S5 gives the knee-fit details, and Supplementary Figure S3 gives the full relative-error curves.

\begin{table}[tp]
\caption{EEGDenoiseNet benchmark comparison. Width-variants base2/base4/base6/base8/base16 of the controlled depthwise-separable U-Net backbone are n=5 results from the fixed statistical aggregation; base10/base12 are n=3 exploratory widths. EEGDN CNN and MicroWaveNet controlled rows were retrained under our pipeline. Literature rows are published reference values and were not retrained. NR: not reported.}
\label{tab:eegdenoisenet}
\centering
\tiny
\setlength{\tabcolsep}{3pt}
\resizebox{\textwidth}{!}{%
\begin{tabular}{lrrrrrrrl}
\hline
Model & Params & EOG CC & EOG T-RRMSE & EOG S-RRMSE & EMG CC & EMG T-RRMSE & EMG S-RRMSE & Source \\
\hline
base2 & 1.05K & \(0.871 \pm 0.014\) & \(0.496 \pm 0.038\) & \(0.474 \pm 0.051\) & \(0.822 \pm 0.007\) & \(0.539 \pm 0.019\) & \(0.353 \pm 0.033\) & controlled \\
base4 & 3.22K & \(0.898 \pm 0.002\) & \(0.426 \pm 0.004\) & \(0.386 \pm 0.009\) & \(0.837 \pm 0.001\) & \(0.503 \pm 0.001\) & \(0.307 \pm 0.004\) & controlled \\
base6 & 6.53K & \(0.904 \pm 0.003\) & \(0.412 \pm 0.009\) & \(0.365 \pm 0.009\) & \(0.840 \pm 0.001\) & \(0.497 \pm 0.003\) & \(0.299 \pm 0.005\) & controlled \\
base8 & 10.99K & \(0.907 \pm 0.001\) & \(0.402 \pm 0.003\) & \(0.353 \pm 0.003\) & \(0.843 \pm 0.001\) & \(0.492 \pm 0.001\) & \(0.292 \pm 0.003\) & controlled \\
base10 & 16.59K & \(0.910 \pm 0.001\) & \(0.396 \pm 0.003\) & \(0.343 \pm 0.005\) & \(0.844 \pm 0.001\) & \(0.490 \pm 0.001\) & \(0.286 \pm 0.001\) & controlled \\
base12 & 23.34K & \(0.912 \pm 0.001\) & \(0.391 \pm 0.002\) & \(0.336 \pm 0.002\) & \(0.844 \pm 0.000\) & \(0.489 \pm 0.000\) & \(0.283 \pm 0.001\) & controlled \\
base16 & 40.26K & \(0.915 \pm 0.001\) & \(0.385 \pm 0.002\) & \(0.320 \pm 0.005\) & \(0.843 \pm 0.002\) & \(0.491 \pm 0.003\) & \(0.285 \pm 0.009\) & controlled \\
EEGDN CNN (controlled) & 8.46M/33.62M & \(0.915 \pm 0.002\) & \(0.374 \pm 0.004\) & \(0.317 \pm 0.001\) & \(0.789 \pm 0.002\) & \(0.606 \pm 0.004\) & \(0.376 \pm 0.006\) & controlled \\
MicroWaveNet (controlled) & 53.50K & \(0.868 \pm 0.052\) & \(0.481 \pm 0.095\) & \(0.432 \pm 0.085\) & \(0.799 \pm 0.003\) & \(0.561 \pm 0.003\) & \(0.365 \pm 0.002\) & controlled \\
EEGDN CNN (published) & 8.40M & 0.923 & 0.336 & 0.343 & 0.786 & 0.632 & 0.604 & literature \\
EEGDN RNN (published) & 787.0K & 0.900 & 0.411 & 0.389 & 0.816 & 0.561 & 0.521 & literature \\
CT-DCENet (published) & 38.10M & 0.947 & 0.246 & NR & 0.933 & 0.304 & NR & literature \\
DeepSeparator (published) & 32.00K & 0.769 & 0.705 & 0.747 & 0.734 & 0.712 & 0.717 & literature \\
MicroWaveNet (published) & 60.00K & 0.910 & 0.378 & 0.404 & 0.830 & 0.467 & 0.389 & literature \\
\hline
\end{tabular}
}
\par\smallskip
{\raggedright\scriptsize Base10 and base12 were evaluated on EEGDenoiseNet only and are omitted from subsequent mixed-artifact, transfer, compute, and downstream analyses for clarity. Rows marked controlled were retrained under our data split, loss, and metric pipeline; rows marked published are point estimates from the original papers and were not rerun. The controlled EEGDN CNN parameter count differs between EOG and EMG because the dense output head scales with segment length. Controlled MicroWaveNet uses our pseudo-Huber loss and pipeline; the original multi-loss Soft-DTW and wavelet-domain recipe was not reproduced. CT-DCENet values are from the original publication; no public code is available for independent verification. \par}
\end{table}

The EEGDenoiseNet results establish the ceiling on the primary single-artifact benchmark; the next question is whether the saturation pattern persists under multi-artifact compositional complexity, harder SNR conditions, and larger test-set scale. The same saturation pattern held on the larger mixed-artifact corpus (Table~\ref{tab:mixed1m}; Figure~\ref{fig:pareto_frontier}B). Base4 reached \(0.728 \pm 0.001\), base8 reached \(0.741 \pm 0.001\), and base16 added a further 0.008 CC at more than three times the parameter count. Supplementary Figure S1 shows the per-seed scatter plots.

\begin{table}[tbp]
\caption{Mixed-1M multi-seed results.}
\label{tab:mixed1m}
\centering
\footnotesize
\begin{tabular}{rrrrrrr}
\hline
Base & Params & CC & RMSE & T-RRMSE & S-RRMSE & SDR \\
\hline
2 & 1.05K & \(0.698 \pm 0.014\) & \(0.233 \pm 0.012\) & \(0.697 \pm 0.024\) & \(0.631 \pm 0.029\) & \(3.59 \pm 0.37\) \\
4 & 3.22K & \(0.728 \pm 0.001\) & \(0.209 \pm 0.001\) & \(0.645 \pm 0.001\) & \(0.573 \pm 0.006\) & \(4.44 \pm 0.02\) \\
6 & 6.53K & \(0.735 \pm 0.002\) & \(0.205 \pm 0.000\) & \(0.633 \pm 0.002\) & \(0.560 \pm 0.002\) & \(4.64 \pm 0.03\) \\
8 & 10.99K & \(0.741 \pm 0.001\) & \(0.202 \pm 0.000\) & \(0.624 \pm 0.001\) & \(0.547 \pm 0.002\) & \(4.79 \pm 0.02\) \\
16 & 40.26K & \(0.749 \pm 0.001\) & \(0.196 \pm 0.000\) & \(0.611 \pm 0.001\) & \(0.534 \pm 0.004\) & \(5.05 \pm 0.01\) \\
\hline
\end{tabular}
\end{table}

To test whether the ceiling extends beyond the training distribution, we evaluated all width variants on BCI Competition IV recordings without fine-tuning. Zero-shot BCI transfer preserved the early-saturation pattern but showed dataset-dependent capacity effects (Table~\ref{tab:bci_zero_shot}). BCI IV-2a improved gradually through base16, whereas BCI IV-2b peaked at base6 and then declined (Figure~\ref{fig:pareto_frontier}C). The IV-2b inversion, base6 \(>\) base8 \(>\) base16, indicates that additional reconstruction capacity can reduce transfer performance.

\begin{table}[tbp]
\caption{BCI zero-shot multi-seed results.}
\label{tab:bci_zero_shot}
\centering
\scriptsize
\resizebox{\textwidth}{!}{%
\begin{tabular}{llrrrrrr}
\hline
Dataset & Base & Params & CC & RMSE & T-RRMSE & S-RRMSE & SDR \\
\hline
IV-2a & 2 & 1.05K & \(0.702 \pm 0.017\) & \(0.471 \pm 0.015\) & \(0.731 \pm 0.023\) & \(0.796 \pm 0.022\) & \(2.81 \pm 0.29\) \\
IV-2a & 4 & 3.22K & \(0.726 \pm 0.004\) & \(0.446 \pm 0.003\) & \(0.693 \pm 0.005\) & \(0.736 \pm 0.013\) & \(3.30 \pm 0.07\) \\
IV-2a & 6 & 6.53K & \(0.731 \pm 0.004\) & \(0.440 \pm 0.003\) & \(0.684 \pm 0.004\) & \(0.712 \pm 0.005\) & \(3.42 \pm 0.06\) \\
IV-2a & 8 & 10.99K & \(0.735 \pm 0.003\) & \(0.437 \pm 0.003\) & \(0.681 \pm 0.004\) & \(0.694 \pm 0.011\) & \(3.47 \pm 0.06\) \\
IV-2a & 16 & 40.26K & \(0.736 \pm 0.004\) & \(0.436 \pm 0.003\) & \(0.678 \pm 0.004\) & \(0.673 \pm 0.014\) & \(3.50 \pm 0.06\) \\
IV-2b & 2 & 1.05K & \(0.750 \pm 0.019\) & \(0.426 \pm 0.020\) & \(0.674 \pm 0.030\) & \(0.545 \pm 0.051\) & \(3.59 \pm 0.41\) \\
IV-2b & 4 & 3.22K & \(0.776 \pm 0.001\) & \(0.398 \pm 0.002\) & \(0.632 \pm 0.003\) & \(0.459 \pm 0.011\) & \(4.18 \pm 0.04\) \\
IV-2b & 6 & 6.53K & \(0.777 \pm 0.001\) & \(0.397 \pm 0.002\) & \(0.630 \pm 0.002\) & \(0.447 \pm 0.002\) & \(4.20 \pm 0.03\) \\
IV-2b & 8 & 10.99K & \(0.772 \pm 0.002\) & \(0.401 \pm 0.002\) & \(0.637 \pm 0.004\) & \(0.451 \pm 0.004\) & \(4.12 \pm 0.05\) \\
IV-2b & 16 & 40.26K & \(0.766 \pm 0.004\) & \(0.408 \pm 0.003\) & \(0.647 \pm 0.004\) & \(0.455 \pm 0.006\) & \(3.98 \pm 0.06\) \\
\hline
\end{tabular}
}
\end{table}

Finally, we examined whether input difficulty or artifact type interacts with the saturation pattern. SNR and artifact stratification show that input difficulty affects absolute CC more than width affects the ceiling (Supplementary Figure S6b, S6c). Width ordering was preserved across SNR bins, with base2 separated most clearly at the lowest SNRs. Artifact type dominated recipe difficulty: EOG-only rows formed a high-CC band, whereas EMG-containing recipes formed a lower-CC band with little left-to-right gradient across widths.

\subsection{Controlled Baselines Confirm the Ceiling Is Not Architecture-Specific}
\label{sec:controlled_baselines}

Controlled same-pipeline retraining shows that the ceiling is not simply an artifact of comparing against weak baselines. EEGDN CNN matched base16 on EOG with CC \(0.915 \pm 0.002\), but required 8.46M parameters for the 512-sample EOG setting. On EMG, EEGDN CNN reached only \(0.789 \pm 0.002\), below all controlled backbone widths (Table~\ref{tab:eegdenoisenet}; Supplementary Table S2).

Controlled MicroWaveNet retraining also fell below the compact operating range of the controlled backbone under the same split, loss, and metric pipeline (Supplementary Table S2). This same-pipeline comparison isolates architectural capacity from training-recipe effects, which is precisely the controlled design this study requires.

A second-family Patch-Transformer sweep showed the same diminishing-return pattern in a non-U-Net architecture. Increasing from 1.35K to 4.87K parameters gave large gains, whereas the 40.71K-parameter variant reached EOG CC \(0.891 \pm 0.001\) and EMG CC \(0.831 \pm 0.001\), below base16 under the same training pipeline (Supplementary Table S4). This control reduces the concern that early saturation is unique to depthwise-separable U-Nets.

The remaining entries in Table~\ref{tab:eegdenoisenet} are literature reference values reported under each method's original evaluation protocol, since publicly available code was not released for most of these models. CT-DCENet reports EOG CC 0.947 with 38.10M parameters \cite{tang2025ctdcenet}, but its evaluation protocol (normalization domain, test-split construction, data augmentation, optimization budget) is not independently verifiable. Even accepting the reported value, a roughly 1000$\times$ parameter increase over base16 yields a 3.2-point CC improvement on EOG -- a marginal gain that reinforces rather than contradicts the diminishing-return pattern. The controlled same-pipeline evidence is more informative: EEGDN CNN at 8.46M parameters matched base16 on EOG, MicroWaveNet did not exceed the compact variants, and the Patch-Transformer showed the same saturation shape. These results support the conclusion that standard benchmarks can be largely saturated by compact temporal models under controlled conditions.

\subsection{Deployment Profiling}
\label{sec:deployment}

\begin{figure}[tbp]
\centering
\includegraphics[width=\textwidth]{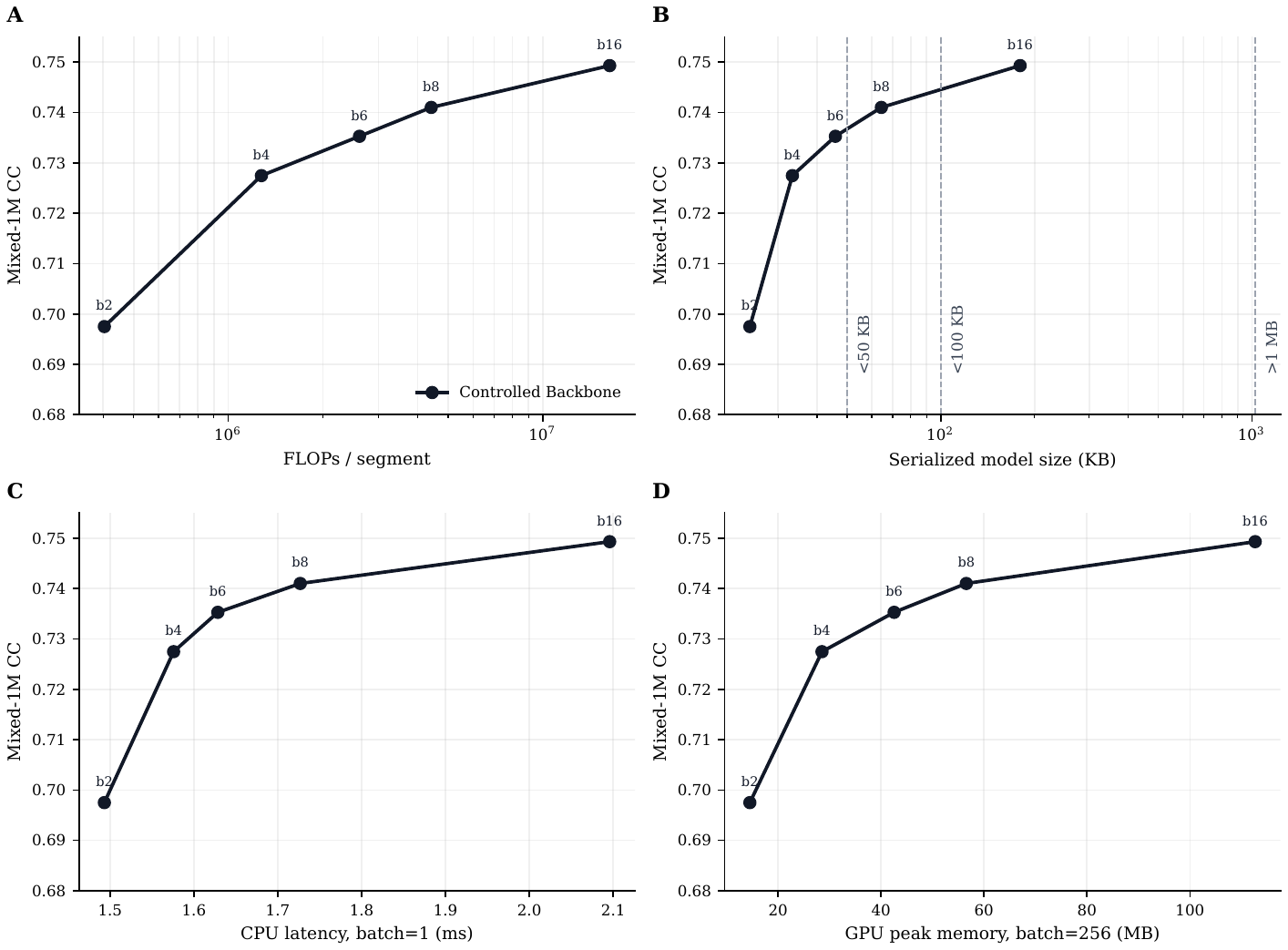}
\caption{Compute efficiency Pareto plots. Panels show CC versus FLOPs, serialized model size, CPU inference latency, and GPU peak memory.}
\label{fig:compute_efficiency}
\end{figure}

The compact Pareto widths occupy deployment-relevant size and compute regimes (Table~\ref{tab:compute}; Figure~\ref{fig:compute_efficiency}). Base4--base6 require 33.3--45.8 KB on disk and 1.27--2.61M FLOPs per 512-sample segment. Base16 improves reconstruction modestly but increases size to 179.8 KB and FLOPs to 16.30M.

\begin{table}[tbp]
\caption{Compute profile on 512-sample inputs. CPU latency is batch-1 mean latency. GPU throughput and peak memory are measured at batch size 256.}
\label{tab:compute}
\centering
\scriptsize
\begin{tabular}{lrrrrrr}
\hline
Model & Params & FLOPs/segment & Size KB & CPU ms & GPU throughput/s & GPU peak KB \\
\hline
base2 & 1.05K & 0.40M & 24.3 & 1.49 & 181\,760 & 14\,881 \\
base4 & 3.22K & 1.27M & 33.3 & 1.58 & 177\,146 & 29\,223 \\
base6 & 6.53K & 2.61M & 45.8 & 1.63 & 165\,138 & 43\,572 \\
base8 & 10.99K & 4.42M & 64.4 & 1.73 & 160\,815 & 57\,924 \\
base16 & 40.26K & 16.30M & 179.8 & 2.10 & 99\,865 & 115\,381 \\
EEGDN CNN & 8.46M & 83.00M & 33\,079 & 2.68 & 34\,249 & 157\,642 \\
EEGDN RNN & 787.0K & 1.58M & 3\,082 & 0.42 & 653\,361 & 34\,443 \\
\hline
\end{tabular}
\end{table}

\subsection{Reconstruction Metrics Do Not Predict Downstream Task Utility}
\label{sec:downstream}

\subsubsection{Baseline Conditions and Experimental Design}

CSP+LDA provided the strongest noisy/noisy baseline among the five downstream decoders (\(0.612 \pm 0.173\)), making it the primary test of whether denoising preserves task-relevant motor-imagery structure (Supplementary Table S6). Because the denoisers are single-channel models applied independently to the 22 EEG channels, the downstream protocol also tests whether high reconstruction fidelity survives the cross-channel covariance requirements of spatial-filter classifiers.

\subsubsection{The Classifier-Dependent Utility Gap}

\begin{figure}[tbp]
\centering
\includegraphics[width=\textwidth]{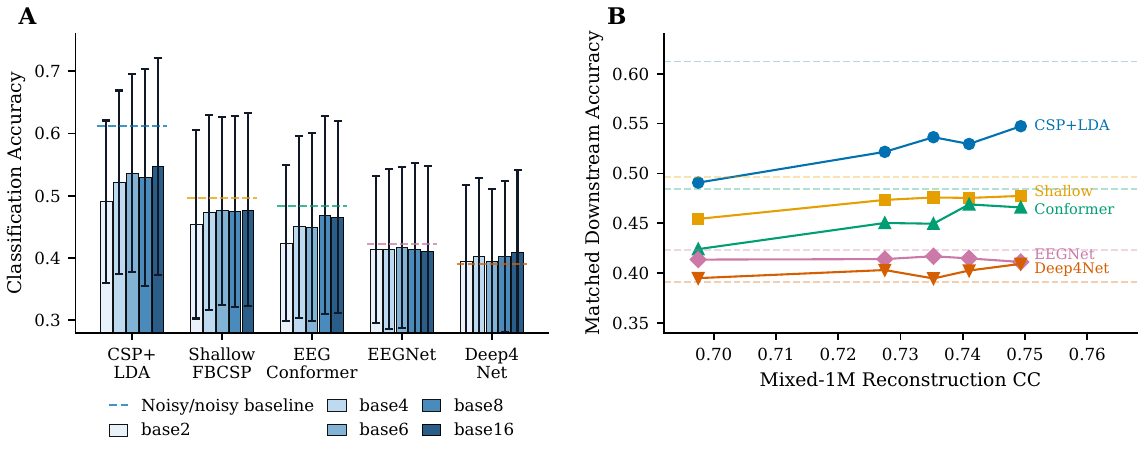}
\caption{Downstream BCI utility gap across classifier families. Panel A shows matched denoised/denoised classification accuracy for each width variant and each classifier, with classifier-specific noisy/noisy baselines indicated by dashed horizontal reference segments. Error bars denote standard deviation across the nine BCI IV-2a subjects, and significance markers above the best fixed width indicate one-sided Wilcoxon tests against noisy/noisy (\(**\): \(p<0.01\), \(*\): \(p<0.05\), n.s.: not significant). Panel B plots Mixed-1M reconstruction CC against matched denoised/denoised downstream accuracy for all width variants and classifiers; dashed horizontal lines show classifier-specific noisy/noisy baselines.}
\label{fig:downstream_gap}
\end{figure}

Denoising harmed spatial-filter pipelines most strongly (Table~\ref{tab:downstream}; Figure~\ref{fig:downstream_gap}). CSP+LDA was below noisy/noisy for all widths, and the best matched denoised/denoised condition remained 0.065 below baseline. This comparison survived correction across the five best-width classifier tests. ShallowFBCSPNet also decreased at its best denoised width, but the result did not survive Bonferroni correction. EEGConformer, EEGNet, and Deep4Net showed non-significant or neutral effects, producing a gradient from spatial-filter sensitivity to end-to-end decoder adaptability. The full width-by-classifier matrix is provided in Supplementary Table S7, per-checkpoint CSP+LDA diagnostics are shown in Supplementary Figure S4, and Figure~\ref{fig:downstream_heatmap} summarizes the corresponding deltas.

\begin{table}[tbp]
\caption{Best matched denoised/denoised downstream BCI IV-2a result for each classifier. Accuracy and deltas are mean \(\pm\) standard deviation across nine subjects. Wilcoxon \(p\)-values compare matched denoised/denoised accuracy against the noisy/noisy baseline with the one-sided alternative that denoised accuracy is lower. Bonferroni and Benjamini–Hochberg false-discovery-rate (BH-FDR) corrections are applied across the five best-width classifier comparisons.}
\label{tab:downstream}
\centering
\footnotesize
\resizebox{\textwidth}{!}{%
\begin{tabular}{lcccccc}
\hline
Classifier & Best fixed width & Accuracy & \(\Delta\) vs noisy/noisy & \(p\) & Bonf. \(p\) & BH \(q\) \\
\hline
CSP+LDA & base16 & \(0.547 \pm 0.174\) & \(-0.065 \pm 0.060\) & 0.0098 & 0.0488 & 0.0488 \\
ShallowFBCSPNet & base16 & \(0.477 \pm 0.155\) & \(-0.019 \pm 0.025\) & 0.0273 & 0.1367 & 0.0684 \\
EEGConformer & base8 & \(0.469 \pm 0.159\) & \(-0.015 \pm 0.045\) & 0.1797 & 0.8984 & 0.2995 \\
EEGNet & base6 & \(0.417 \pm 0.129\) & \(-0.006 \pm 0.056\) & 0.2480 & 1.0000 & 0.3101 \\
Deep4Net & base16 & \(0.409 \pm 0.132\) & \(+0.018 \pm 0.050\) & 0.8496 & 1.0000 & 0.8496 \\
\hline
\end{tabular}
}
\end{table}

\begin{figure}[tbp]
\centering
\includegraphics[width=0.65\textwidth]{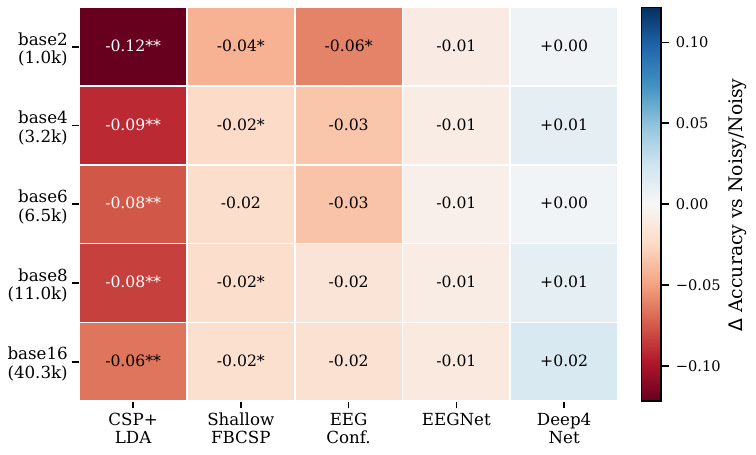}
\caption{Width \(\times\) classifier utility-gap heatmap. Each cell shows the change in matched denoised/denoised classification accuracy relative to the noisy/noisy baseline, averaged across all nine BCI IV-2a subjects. Negative values indicate denoising-induced degradation; positive values indicate improvement. Significance markers denote one-sided Wilcoxon signed-rank tests against noisy/noisy (\(**\): \(p<0.01\), \(*\): \(p<0.05\), uncorrected).}
\label{fig:downstream_heatmap}
\end{figure}

\subsubsection{The Gap Is Not Artifact-Specific}

The CSP+LDA gap persisted across artifact recipes (Table~\ref{tab:contamination_robustness}). The EOG row anchors this analysis to the main synthetic-EOG CSP+LDA results from Table~\ref{tab:downstream}, while EMG and mixed EOG+EMG+LINE contamination produced larger reduction. All nine subjects were below the noisy/noisy baseline at the best fixed width for EMG and mixed contamination.

\begin{table}[tbp]
\caption{CSP+LDA downstream robustness across synthetic artifact recipes. Values are mean \(\pm\) standard deviation across nine BCI IV-2a subjects. Wilcoxon \(p\)-values compare the best matched denoised/denoised condition against noisy/noisy with the one-sided alternative that denoised accuracy is lower.}
\label{tab:contamination_robustness}
\centering
\footnotesize
\resizebox{\textwidth}{!}{%
\begin{tabular}{lccccc}
\hline
Contaminant & Noisy/noisy & Best width & Best denoised/denoised & \(\Delta\) vs noisy/noisy & \(p\) / BH \(q\) \\
\hline
EOG & \(0.612 \pm 0.173\) & base16 & \(0.547 \pm 0.174\) & \(-0.065 \pm 0.060\) & 0.0098 / 0.0098 \\
EMG & \(0.613 \pm 0.164\) & base6 & \(0.504 \pm 0.169\) & \(-0.109 \pm 0.080\) & 0.0020 / 0.0059 \\
EOG+EMG+LINE & \(0.610 \pm 0.166\) & base4 & \(0.497 \pm 0.163\) & \(-0.114 \pm 0.086\) & 0.0020 / 0.0059 \\
\hline
\end{tabular}
}
\end{table}

\subsubsection{The Gap Persists on Real Recordings}

The utility gap also appeared when denoisers were applied directly to naturally recorded BCI IV-2a trials without synthetic artifact injection (Table~\ref{tab:real_bci_downstream}). The best matched denoised CSP+LDA condition was base8 at \(0.597 \pm 0.178\), below raw/raw by 4.7 percentage points, and all matched denoised widths were significant after BH-FDR correction. This result shows that the spatial-filter gap is not an artifact of the synthetic contamination protocol.

\begin{table}[tbp]
\caption{Real non-synthetic BCI IV-2a CSP+LDA evaluation. Denoisers were applied directly to recorded trials without synthetic EOG injection. Values are mean \(\pm\) standard deviation across nine subjects. BH-FDR \(q\)-values correct the matched denoised/denoised comparisons across widths.}
\label{tab:real_bci_downstream}
\centering
\footnotesize
\begin{tabular}{lcccc}
\hline
Condition & Width & Accuracy & \(\Delta\) vs raw/raw & BH-FDR \(q\) \\
\hline
raw/raw & -- & \(0.643 \pm 0.160\) & -- & -- \\
denoised/denoised & base2 & \(0.544 \pm 0.146\) & \(-0.100 \pm 0.051\) & 0.0049 \\
denoised/denoised & base4 & \(0.583 \pm 0.166\) & \(-0.060 \pm 0.040\) & 0.0049 \\
denoised/denoised & base6 & \(0.593 \pm 0.180\) & \(-0.051 \pm 0.052\) & 0.0059 \\
denoised/denoised & base8 & \(0.597 \pm 0.178\) & \(-0.047 \pm 0.038\) & 0.0049 \\
denoised/denoised & base16 & \(0.594 \pm 0.183\) & \(-0.049 \pm 0.044\) & 0.0049 \\
\hline
\end{tabular}
\end{table}

\subsubsection{The Gap Is Not Explained by Independent-Channel Processing}

Because the controlled backbone is a single-channel denoiser applied independently to 22 EEG channels, it may distort the cross-channel covariance structure that CSP+LDA depends on. To isolate this confound, we trained subject-specific 22-channel DSConv U-Net denoisers at base6 and base16 on the BCI IV-2a training sessions and evaluated matched CSP+LDA on the held-out evaluation sessions (Table~\ref{tab:multichannel_control}).

\begin{table}[tbp]
\caption{Multichannel denoiser control for CSP+LDA downstream evaluation. Single-channel rows apply the denoiser independently to each EEG channel. Multichannel control rows use subject-specific 22-channel DSConv U-Net denoisers trained on the BCI IV-2a training session. Values are mean \(\pm\) standard deviation across nine subjects.}
\label{tab:multichannel_control}
\centering
\footnotesize
\resizebox{\textwidth}{!}{%
\begin{tabular}{lcccccc}
\hline
Scope & Width & Params & Accuracy & \(\Delta\) vs noisy/noisy & Subjects below & \(p\) / BH \(q\) \\
\hline
Noisy/noisy baseline & -- & -- & \(0.612 \pm 0.173\) & -- & -- & -- \\
Single-channel & base6 & 6.53K & \(0.536 \pm 0.159\) & \(-0.076 \pm 0.068\) & 7/9 & 0.0098 / 0.0130 \\
Single-channel & base16 & 40.26K & \(0.547 \pm 0.174\) & \(-0.065 \pm 0.060\) & 7/9 & 0.0098 / 0.0130 \\
Multichannel control & base6 & 88.87K & \(0.513 \pm 0.163\) & \(-0.100 \pm 0.078\) & 7/9 & 0.0098 / 0.0130 \\
Multichannel control & base16 & 602.99K & \(0.567 \pm 0.174\) & \(-0.045 \pm 0.076\) & 6/9 & 0.0645 / 0.0645 \\
\hline
\end{tabular}
}
\end{table}

Multichannel base16 reduced the gap relative to single-channel denoising (\(\Delta=-0.045\) versus \(-0.065\)), but multichannel base6 produced a larger reduction (\(\Delta=-0.100\)). Even the best multichannel condition did not reach the noisy/noisy baseline. Covariance-aware denoising partially mitigates but does not eliminate the utility gap, confirming that both amplitude suppression and cross-channel distortion contribute independently to the downstream degradation.

\subsubsection{Calibration and Convergence Controls}

Post-denoising train-channel z-scoring, raw-variance rescaling, and covariance recoloring did not recover noisy/noisy performance; per-trial z-scoring reduced the average gap but remained below baseline (Supplementary Table S8). A 200-epoch convergence audit improved Deep4Net and EEGConformer noisy/noisy baselines, but their denoising effects remained statistically inconclusive (Supplementary Table S9). Supplementary Tables S6--S10 provide the full downstream archive, including baseline, width-matrix, calibration, convergence, and task-aware pilot summaries. These controls indicate that the utility gap is not explained by simple calibration failure or neural-decoder under-training.

\subsubsection{Qualitative Mechanism: Amplitude Suppression in Quiet Intervals}

Real-artifact examples provide a mechanism for the downstream results (Supplementary Figure S7). In high-EOG windows, all widths removed most of the blink-scale signal component. In quiet-baseline windows, however, base4--base16 still removed 35--59\% of raw RMS depending on dataset, even when visible artifact activity was low. This over-stripping supports an amplitude-suppression mechanism: semi-synthetic reconstruction training can teach the model to treat high-amplitude structure as artifact even when that structure may carry task-relevant neural information. Supplementary Figure S2 shows high-amplitude transient examples, and Supplementary Figure S5 gives the full removed/raw RMS heatmap.

\section{Discussion}

\subsection{Why the Ceiling Exists}

The width sweep shows that these benchmarks saturate with compact temporal models. Most reconstruction gain appears below 10K parameters, and the EMG post-elbow slope is statistically indistinguishable from zero under the descriptive bootstrap analysis. Performance continues to improve marginally at higher widths, but the benchmark mapping becomes easy relative to the capacity added beyond base4--base8, consistent with a low-dimensional separation problem.

A plausible explanation is low-dimensional spectral separation. EOG artifacts are dominated by low-frequency structure, EMG is broadband and overlaps more strongly with neural activity, and line noise is narrowband. The artifact-stratified heatmap supports this mechanism: EMG presence changes difficulty far more than width, while line noise and ECG additions have smaller effects once EMG is present. A compact temporal convolutional model with dilation can learn these coarse temporal and spectral boundaries without requiring millions of parameters.

The BCI IV-2b inversion reinforces this interpretation. Larger models did not produce more transferable denoising; base6 exceeded base16 on that zero-shot transfer setting. Extra capacity can therefore fit training-distribution details rather than learning a more general artifact-removal rule. The Patch-Transformer control shows that early diminishing returns are not unique to the DSConv U-Net template. Uncontrolled literature-reported scores from models without publicly available code cannot isolate whether apparent gains reflect architectural capacity, training-recipe details, or evaluation-protocol differences, and therefore do not contradict the controlled same-pipeline evidence.

\subsection{Why Metrics and Utility Diverge}

Reconstruction metrics and downstream utility diverge because reconstruction training can remove signal components that classifiers use. As shown in the qualitative analysis (Supplementary Figure S7), denoisers removed substantial signal energy even in quiet-baseline windows, consistent with an amplitude-suppression bias from semi-synthetic training.

The single-channel-to-multichannel interface adds a second mechanism. The controlled backbone denoises each EEG channel independently before spatial-filter classification, while CSP+LDA depends on cross-channel covariance and band-power relationships. However, the controls show that covariance distortion alone is not sufficient to explain the gap. Post-denoising z-scoring, raw-variance rescaling, and covariance recoloring did not recover noisy/noisy performance, and a focused 22-channel multichannel denoiser reduced but did not eliminate the CSP+LDA reduction (Table~\ref{tab:multichannel_control}). The utility loss therefore reflects both amplitude suppression and task-relevant representational change.

Classifier architecture determines how strongly this mismatch appears. CSP+LDA is linear and cannot learn compensatory filters after denoising; ShallowFBCSPNet inherits similar band-power sensitivity. End-to-end decoders can partly adapt to the altered signal distribution, which explains why EEGNet, Deep4Net, and EEGConformer showed variable or neutral effects. The 200-epoch audit improved neural-decoder baselines but did not convert denoising into a reliable downstream benefit.

A task-aware pilot tested whether adding frozen CSP-feature preservation to a 22-channel denoiser could close the gap. It did not: the auxiliary loss worsened matched CSP+LDA performance relative to reconstruction-only multichannel denoising, with \(\Delta=-0.061\) at \(\lambda=0.1\) and \(\Delta=-0.102\) at \(\lambda=1.0\), compared with \(\Delta=-0.043\) for reconstruction-only training (Supplementary Table S10). This result demonstrates that closing the metric--utility gap requires more principled task-aware optimization than simple noisy-feature matching, and identifies frozen-feature preservation as an insufficient approach.

The interpretation should be read alongside the standard assumption of supervised EEG denoising: the ``clean'' reference is itself a preprocessed approximation, not an independently observed artifact-free neural source \cite{delorme2004eeglab,bigdely2015prep,nolan2010faster,mognon2011adjust,urigueen2015artifact}. This assumption is shared by every method evaluated on EEGDenoiseNet and related benchmarks. Our results therefore characterize the capacity requirements of the evaluation standard the field currently uses, which is the necessary first step toward designing harder benchmarks that better approximate real artifact removal.

\subsection{Implications and Recommendations}

For practitioners deploying edge EEG, base4--base6 are the practical operating points in this study. They use 3.22K--6.53K parameters, occupy 33--46 KB on disk, and preserve most of the reconstruction performance of larger variants. Million-parameter denoisers should therefore be justified by application-specific improvements in robustness, transfer, or task utility, not by reconstruction metrics alone.

For researchers publishing denoising methods, downstream task evaluation should accompany CC, RMSE, SDR, and spectral-error metrics. The present results show that a denoiser can improve reconstruction while degrading CSP+LDA classification across subjects, artifact types, and naturally recorded trials. Reconstruction fidelity remains necessary, but it should be treated as an intermediate diagnostic rather than a complete success criterion.

For benchmark designers, the next step is to move beyond semi-synthetic clean-reference mixtures alone. The Temple University Hospital EEG corpus provides clinical recordings with real, heterogeneous artifacts \cite{obeid2016temple}, but it lacks paired clean targets. Future benchmarks should combine realistic artifact conditions with task-aware endpoints such as motor-imagery decoding, ERP detection, seizure-marker preservation, or spectral biomarker stability.

For loss-function designers, task-aware denoising is promising but technically delicate. Recent task-oriented denoising proposals point in this direction \cite{xiang2025taskoriented}, but our CSP-feature pilot shows that naive feature preservation can worsen downstream performance. Three directions appear more viable than frozen-feature matching. First, differentiable end-to-end pipelines that backpropagate classifier gradients through the denoiser could jointly optimize reconstruction fidelity and task utility, avoiding the proxy mismatch of separately trained objectives. Second, band-selective or spectrally constrained losses that explicitly penalize attenuation in task-relevant frequency bands (e.g., 8--30 Hz mu/beta for motor imagery, 1--4 Hz delta for sleep staging) could reduce neural feature suppression without requiring a full downstream classifier during training. Third, multi-objective Pareto training that treats reconstruction and downstream preservation as competing objectives could identify operating points that balance fidelity and utility rather than forcing a single-loss compromise.

Based on the present findings, we recommend that future EEG denoising studies include a minimal downstream evaluation protocol: (i) report matched denoised/denoised classification accuracy on at least one established BCI or clinical paradigm alongside reconstruction metrics; (ii) include both a linear spatial-filter decoder (e.g., CSP+LDA) and at least one end-to-end neural decoder to expose classifier-dependent effects; and (iii) test on naturally recorded trials without synthetic artifact injection to confirm that denoiser effects generalize beyond the semi-synthetic training distribution.

\subsection{Limitations}

This study characterizes the capacity requirements and downstream utility of the dominant supervised EEG denoising evaluation paradigm; it does not claim that all real-world denoising tasks are equally low-dimensional. The capacity sweep was conducted on single-channel inputs, consistent with the EEGDenoiseNet benchmark protocol that serves as the primary evaluation standard in the field. A focused 22-channel multichannel denoiser control confirmed that the CSP+LDA downstream gap is not an artifact of independent per-channel processing (Table~\ref{tab:multichannel_control}), but multichannel reconstruction benchmarking and deployment-optimized multichannel architectures are beyond the scope of this study.

The primary reconstruction benchmarks are semi-synthetic, which is the standard evaluation paradigm for supervised EEG denoising. To test whether conclusions transfer beyond semi-synthetic conditions, we evaluated denoisers on naturally recorded BCI IV-2a trials without synthetic artifact injection (Table~\ref{tab:real_bci_downstream}); the CSP+LDA utility gap persisted. Downstream evaluation focused on motor imagery across all nine BCI IV-2a subjects and five decoder families rather than spanning multiple BCI paradigms or clinical tasks. This within-paradigm depth with nine subjects, five classifiers, three artifact types, and real and synthetic conditions prioritizes statistical rigor over paradigm breadth.

The controlled same-pipeline baseline comparisons cover EEGDN CNN, MicroWaveNet, and a Patch-Transformer capacity sweep as a second-family control. CT-DCENet could not be independently verified because the official code was unavailable; its literature-reported values are retained as uncontrolled reference points and labeled as such. The Patch-Transformer control confirms that the diminishing-return pattern is not specific to depthwise-separable U-Nets, and the convergence of both families toward the same saturation pattern supports generality beyond a single architecture template.

\section{Conclusion}

Standard EEG denoising benchmarks saturate with compact models: 3--6.5K parameters recover most reconstruction performance, and the pattern holds across the primary DSConv family, controlled baseline retraining, and a second-family Patch-Transformer control. Same-pipeline evidence shows that benchmark difficulty is far below the million-parameter regime, and that the marginal reconstruction gains reported by larger models come at parameter costs of 200--1000$\times$ with no demonstrated downstream benefit.

Reconstruction-optimized denoising degrades spatial-filter motor-imagery classification across artifact types and on real non-synthetic recordings, while end-to-end neural decoders show variable effects. Reconstruction fidelity is therefore necessary but not sufficient for judging EEG denoising utility.

For edge EEG, base4--base6 models at 33--46 KB are the recommended operating points from this study. EEG denoising should be evaluated not only by fidelity to a reconstructed reference, but by whether it preserves the neural information that downstream applications require. We recommend that denoising papers report matched classifier accuracy on at least one BCI or clinical paradigm, using both linear and end-to-end decoders, to detect the classifier-dependent utility effects that reconstruction metrics alone cannot capture.

\section*{Author Contributions}
J.S.B. conceived the idea, designed the experiments, and implemented them. S.P. supervised the research. All authors reviewed and approved the final manuscript.

\section*{Conflicts of Interest}
The authors declare no conflicts of interest.

\funding{This research received no specific grant from any funding agency in the public, commercial, or not-for-profit sectors.}

\data{The EEGDenoiseNet benchmark dataset is publicly available \cite{zhang2021eegdenoisenet}. BCI Competition IV datasets are available from the competition website. The Mixed-1M corpus generation code, model training code, evaluation scripts, and checkpoint generation will be made available pon publication of the peer-reviewed version of this work.}

\section*{Ethical Statement}
This study used exclusively publicly available benchmark datasets (EEGDenoiseNet and BCI Competition IV). These datasets were collected and published under ethical approvals obtained in the original studies. No new human or animal data were collected for this work.

\section*{References}

\bibliographystyle{iopart-num}
\bibliography{references}

@article{zhang2021eegdenoisenet,
  author  = {Zhang, Haoming and Zhao, Mingqi and Wei, Chen and Mantini, Dante and Li, Zherui and Liu, Quanying},
  title   = {{EEGdenoiseNet}: A Benchmark Dataset for Deep Learning Solutions of {EEG} Denoising},
  journal = {Journal of Neural Engineering},
  year    = {2021},
  volume  = {18},
  number  = {5},
  pages   = {056057},
  doi     = {10.1088/1741-2552/ac2bf8}
}

@article{obeid2016temple,
  title   = {The Temple University Hospital {EEG} Data Corpus},
  author  = {Obeid, Iyad and Picone, Joseph},
  journal = {Frontiers in Neuroscience},
  volume  = {10},
  pages   = {196},
  year    = {2016},
  doi     = {10.3389/fnins.2016.00196}
}

@article{chuang2022icunet,
  author  = {Chuang, Chun-Hsiang and Chang, Kong-Yi and Huang, Chih-Sheng and Jung, Tzyy-Ping},
  title   = {{IC-U-Net}: A {U-Net}-Based Denoising Autoencoder Using Mixtures of Independent Components for Automatic {EEG} Artifact Removal},
  journal = {NeuroImage},
  year    = {2022},
  volume  = {263},
  pages   = {119586},
  doi     = {10.1016/j.neuroimage.2022.119586}
}

@article{yue2025lrrunet,
  author  = {Yue, Xiaoxiong and Lu, Liangfu and Liu, Haipeng and Zang, Yunliang},
  title   = {{LRR-UNet}: A Deep Unfolding Network With Low-Rank Recovery for {EEG} Signal Denoising},
  journal = {CNS Neuroscience \& Therapeutics},
  year    = {2025},
  volume  = {31},
  number  = {10},
  pages   = {e70632},
  doi     = {10.1111/cns.70632}
}

@article{dong2023wgan,
  author  = {Dong, Yuanzhe and Tang, Xi and Li, Qingge and Wang, Yingying and Jiang, Naifu and Tian, Lan and Zheng, Yue and Li, Xiangxin and Zhao, Shaofeng and Li, Guanglin and Fang, Peng},
  title   = {An Approach for {EEG} Denoising Based on Wasserstein Generative Adversarial Network},
  journal = {IEEE Transactions on Neural Systems and Rehabilitation Engineering},
  year    = {2023},
  volume  = {31},
  pages   = {3524--3534},
  doi     = {10.1109/TNSRE.2023.3309815}
}

@article{gao2023duocl,
  author  = {Gao, Tengfei and Chen, Dan and Tang, Yunbo and Ming, Zhekai and Li, Xiaoli},
  title   = {{EEG} Reconstruction With a Dual-Scale {CNN-LSTM} Model for Deep Artifact Removal},
  journal = {IEEE Journal of Biomedical and Health Informatics},
  year    = {2023},
  volume  = {27},
  number  = {3},
  pages   = {1283--1294},
  doi     = {10.1109/JBHI.2022.3227320}
}

@article{yin2025gctnet,
  author  = {Yin, Jin and Liu, Aiping and Li, Chang and Qian, Ruobing and Chen, Xun},
  title   = {A {GAN} Guided Parallel {CNN} and Transformer Network for {EEG} Denoising},
  journal = {IEEE Journal of Biomedical and Health Informatics},
  year    = {2025},
  volume  = {29},
  number  = {6},
  pages   = {3930--3941},
  doi     = {10.1109/JBHI.2023.3277596}
}

@article{chen2024denosieformer,
  author  = {Chen, Junfu and Pi, Dechang and Jiang, Xiaoyi and Xu, Yue and Chen, Yang and Wang, Xixuan},
  title   = {Denosieformer: A Transformer Based Approach for Single-Channel {EEG} Artifact Removal},
  journal = {IEEE Transactions on Instrumentation and Measurement},
  year    = {2024},
  volume  = {73},
  pages   = {1--16},
  doi     = {10.1109/TIM.2023.3341114}
}

@article{tang2025ctdcenet,
  author  = {Tang, Yunbo and Huang, Weirong and Chen, Chuanxi and Chen, Dan},
  title   = {{CT-DCENet}: Deep {EEG} Denoising via {CNN-Transformer}-Based Dual-Stage Collaborative Ensemble Learning},
  journal = {IEEE Journal of Biomedical and Health Informatics},
  year    = {2025},
  volume  = {29},
  number  = {6},
  pages   = {4095--4108},
  doi     = {10.1109/JBHI.2025.3535592}
}

@article{chen2025denoisemamba,
  author  = {Chen, Wensheng and Li, Yurong and Zheng, Nan and Shi, Wuxiang},
  title   = {{DenoiseMamba}: An Innovative Approach for {EEG} Artifact Removal Leveraging {Mamba} and {CNN}},
  journal = {IEEE Journal of Biomedical and Health Informatics},
  year    = {2025},
  volume  = {29},
  number  = {9},
  pages   = {6551--6564},
  doi     = {10.1109/JBHI.2025.3573042}
}

@article{yu2022deepseparator,
  author  = {Yu, Junjie and Li, Chenyi and Lou, Kexin and Wei, Chen and Liu, Quanying},
  title   = {Embedding Decomposition for Artifacts Removal in {EEG} Signals},
  journal = {Journal of Neural Engineering},
  year    = {2022},
  volume  = {19},
  number  = {2},
  pages   = {026052},
  doi     = {10.1088/1741-2552/ac63eb}
}

@inproceedings{lahiri2025microwavenet,
  author    = {Lahiri, Jeet Bandhu and Kulkarni, Arvasu and Panwar, Siddharth},
  title     = {{MicroWaveNet}: Lightweight {CBAM}-Augmented Wavelet-Attentive Networks for Robust {EEG} Denoising},
  booktitle = {2025 IEEE 35th International Workshop on Machine Learning for Signal Processing (MLSP)},
  year      = {2025},
  pages     = {1--6},
  doi       = {10.1109/MLSP62443.2025.11204235}
}

@article{huang2024ltdnet,
  author  = {Huang, Jingwei and Wang, Chuansheng and Zhao, Wanqi and Grau, Antoni and Xue, Xingsi and Zhang, Fuquan},
  title   = {{LTDNet-EEG}: A Lightweight Network of Portable/Wearable Devices for Real-Time {EEG} Signal Denoising},
  journal = {IEEE Transactions on Consumer Electronics},
  year    = {2024},
  volume  = {70},
  number  = {3},
  pages   = {5561--5575},
  doi     = {10.1109/TCE.2024.3412774}
}

@article{lawhern2018eegnet,
  author  = {Lawhern, Vernon J. and Solon, Amelia J. and Waytowich, Nicholas R. and Gordon, Stephen M. and Hung, Chou P. and Lance, Brent J.},
  title   = {{EEGNet}: A Compact Convolutional Neural Network for {EEG}-Based Brain--Computer Interfaces},
  journal = {Journal of Neural Engineering},
  year    = {2018},
  volume  = {15},
  number  = {5},
  pages   = {056013},
  doi     = {10.1088/1741-2552/aace8c}
}

@article{nagar2021compressed,
  author  = {Nagar, Subham and Kumar, Ahlad and Swamy, M. N. S.},
  title   = {Orthogonal Features-Based {EEG} Signal Denoising Using Fractionally Compressed Autoencoder},
  journal = {Signal Processing},
  year    = {2021},
  volume  = {188},
  pages   = {108225},
  doi     = {10.1016/j.sigpro.2021.108225}
}

@article{tangermann2012bci,
  author  = {Tangermann, Michael and M{\"u}ller, Klaus-Robert and Aertsen, Ad and Birbaumer, Niels and Braun, Christoph and Brunner, Clemens and Leeb, Robert and Mehring, Carsten and Miller, Kai J. and M{\"u}ller-Putz, Gernot R. and Nolte, Guido and Pfurtscheller, Gert and Preissl, Hubert and Schalk, Gerwin and Schl{\"o}gl, Alois and Vidaurre, Carmen and Waldert, Stephan and Blankertz, Benjamin},
  title   = {Review of the {BCI} Competition {IV}},
  journal = {Frontiers in Neuroscience},
  year    = {2012},
  volume  = {6},
  pages   = {55},
  doi     = {10.3389/fnins.2012.00055}
}

@article{ramoser2000csp,
  author  = {Ramoser, Herbert and M{\"u}ller-Gerking, Johannes and Pfurtscheller, Gert},
  title   = {Optimal Spatial Filtering of Single Trial {EEG} During Imagined Hand Movement},
  journal = {IEEE Transactions on Rehabilitation Engineering},
  year    = {2000},
  volume  = {8},
  number  = {4},
  pages   = {441--446},
  doi     = {10.1109/86.895946}
}

@article{croft2000ocular,
  author  = {Croft, Rodney J. and Barry, Robert J.},
  title   = {Removal of Ocular Artifact from the {EEG}: A Review},
  journal = {Neurophysiologie Clinique/Clinical Neurophysiology},
  year    = {2000},
  volume  = {30},
  number  = {1},
  pages   = {5--19},
  doi     = {10.1016/S0987-7053(00)00055-1}
}

@article{gratton1983ocular,
  author  = {Gratton, Gabriele and Coles, Michael G. H. and Donchin, Emanuel},
  title   = {A New Method for Off-Line Removal of Ocular Artifact},
  journal = {Electroencephalography and Clinical Neurophysiology},
  year    = {1983},
  volume  = {55},
  number  = {4},
  pages   = {468--484},
  doi     = {10.1016/0013-4694(83)90135-9}
}

@article{jung2000bss,
  author  = {Jung, Tzyy-Ping and Makeig, Scott and Humphries, Colin and Lee, Te-Won and McKeown, Martin J. and Iragui, Vicente and Sejnowski, Terrence J.},
  title   = {Removing Electroencephalographic Artifacts by Blind Source Separation},
  journal = {Psychophysiology},
  year    = {2000},
  volume  = {37},
  number  = {2},
  pages   = {163--178},
  doi     = {10.1111/1469-8986.3720163}
}

@article{urigueen2015artifact,
  author  = {Urig{\"u}en, Jose A. and Garcia-Zapirain, Bego{\~n}a},
  title   = {{EEG} Artifact Removal-State-of-the-Art and Guidelines},
  journal = {Journal of Neural Engineering},
  year    = {2015},
  volume  = {12},
  number  = {3},
  pages   = {031001},
  doi     = {10.1088/1741-2560/12/3/031001}
}

@inproceedings{makeig1996ica,
  author    = {Makeig, Scott and Bell, Anthony J. and Jung, Tzyy-Ping and Sejnowski, Terrence J.},
  title     = {Independent Component Analysis of Electroencephalographic Data},
  booktitle = {Advances in Neural Information Processing Systems},
  year      = {1996},
  volume    = {8},
  pages     = {145--151}
}

@article{delorme2004eeglab,
  author  = {Delorme, Arnaud and Makeig, Scott},
  title   = {{EEGLAB}: An Open Source Toolbox for Analysis of Single-Trial {EEG} Dynamics Including Independent Component Analysis},
  journal = {Journal of Neuroscience Methods},
  year    = {2004},
  volume  = {134},
  number  = {1},
  pages   = {9--21},
  doi     = {10.1016/j.jneumeth.2003.10.009}
}

@article{nolan2010faster,
  author  = {Nolan, Hugh and Whelan, Robert and Reilly, Richard B.},
  title   = {{FASTER}: Fully Automated Statistical Thresholding for {EEG} Artifact Rejection},
  journal = {Journal of Neuroscience Methods},
  year    = {2010},
  volume  = {192},
  number  = {1},
  pages   = {152--162},
  doi     = {10.1016/j.jneumeth.2010.07.015}
}

@article{mullen2015realtime,
  author  = {Mullen, Tim R. and Kothe, Christian A. E. and Chi, Yu Mike and Ojeda, Alejandro and Kerth, Trevor and Makeig, Scott and Jung, Tzyy-Ping and Cauwenberghs, Gert},
  title   = {Real-Time Neuroimaging and Cognitive Monitoring Using Wearable Dry {EEG}},
  journal = {IEEE Transactions on Biomedical Engineering},
  year    = {2015},
  volume  = {62},
  number  = {11},
  pages   = {2553--2567},
  doi     = {10.1109/TBME.2015.2481482}
}

@article{chang2020asr,
  author  = {Chang, Chia-Yuan and Hsu, Sheng-Hsiou and Pion-Tonachini, Luca and Jung, Tzyy-Ping},
  title   = {Evaluation of Artifact Subspace Reconstruction for Automatic Artifact Components Removal in Multi-Channel {EEG} Recordings},
  journal = {IEEE Transactions on Biomedical Engineering},
  year    = {2020},
  volume  = {67},
  number  = {4},
  pages   = {1114--1121},
  doi     = {10.1109/TBME.2019.2930186}
}

@article{bigdely2015prep,
  author  = {Bigdely-Shamlo, Nima and Mullen, Tim and Kothe, Christian and Su, Kyung-Min and Robbins, Kay A.},
  title   = {The {PREP} Pipeline: Standardized Preprocessing for Large-Scale {EEG} Analysis},
  journal = {Frontiers in Neuroinformatics},
  year    = {2015},
  volume  = {9},
  pages   = {16},
  doi     = {10.3389/fninf.2015.00016}
}

@article{mognon2011adjust,
  author  = {Mognon, Andrea and Jovicich, Jorge and Bruzzone, Lorenzo and Buiatti, Marco},
  title   = {{ADJUST}: An Automatic {EEG} Artifact Detector Based on the Joint Use of Spatial and Temporal Features},
  journal = {Psychophysiology},
  year    = {2011},
  volume  = {48},
  number  = {2},
  pages   = {229--240},
  doi     = {10.1111/j.1469-8986.2010.01061.x}
}

@article{parra2005recipes,
  author  = {Parra, Lucas C. and Spence, Clay D. and Gerson, Adam D. and Sajda, Paul},
  title   = {Recipes for the Linear Analysis of {EEG}},
  journal = {NeuroImage},
  year    = {2005},
  volume  = {28},
  number  = {2},
  pages   = {326--341},
  doi     = {10.1016/j.neuroimage.2005.05.032}
}

@inproceedings{chollet2017xception,
  author    = {Chollet, Fran{\c{c}}ois},
  title     = {{Xception}: Deep Learning with Depthwise Separable Convolutions},
  booktitle = {Proceedings of the IEEE Conference on Computer Vision and Pattern Recognition},
  year      = {2017},
  pages     = {1251--1258},
  doi       = {10.1109/CVPR.2017.195}
}

@article{howard2017mobilenets,
  author  = {Howard, Andrew G. and Zhu, Menglong and Chen, Bo and Kalenichenko, Dmitry and Wang, Weijun and Weyand, Tobias and Andreetto, Marco and Adam, Hartwig},
  title   = {{MobileNets}: Efficient Convolutional Neural Networks for Mobile Vision Applications},
  journal = {arXiv preprint arXiv:1704.04861},
  year    = {2017},
  eprint  = {1704.04861},
  archivePrefix = {arXiv},
  primaryClass  = {cs.CV}
}

@inproceedings{wang2020ecanet,
  author    = {Wang, Qilong and Wu, Banggu and Zhu, Pengfei and Li, Peihua and Zuo, Wangmeng and Hu, Qinghua},
  title     = {{ECA-Net}: Efficient Channel Attention for Deep Convolutional Neural Networks},
  booktitle = {Proceedings of the IEEE/CVF Conference on Computer Vision and Pattern Recognition},
  year      = {2020},
  pages     = {11531--11539},
  doi       = {10.1109/CVPR42600.2020.01155}
}

@inproceedings{han2016deepcompression,
  author    = {Han, Song and Mao, Huizi and Dally, William J.},
  title     = {Deep Compression: Compressing Deep Neural Networks with Pruning, Trained Quantization and Huffman Coding},
  booktitle = {International Conference on Learning Representations},
  year      = {2016}
}

@article{pfurtscheller2001motor,
  author  = {Pfurtscheller, Gert and Neuper, Christa},
  title   = {Motor Imagery and Direct Brain-Computer Communication},
  journal = {Proceedings of the IEEE},
  year    = {2001},
  volume  = {89},
  number  = {7},
  pages   = {1123--1134},
  doi     = {10.1109/5.939829}
}

@article{blankertz2008spatial,
  author  = {Blankertz, Benjamin and Tomioka, Ryota and Lemm, Steven and Kawanabe, Motoaki and M{\"u}ller, Klaus-Robert},
  title   = {Optimizing Spatial Filters for Robust {EEG} Single-Trial Analysis},
  journal = {IEEE Signal Processing Magazine},
  year    = {2008},
  volume  = {25},
  number  = {1},
  pages   = {41--56},
  doi     = {10.1109/MSP.2008.4408441}
}

@article{lotte2018review,
  author  = {Lotte, Fabien and Bougrain, Laurent and Cichocki, Andrzej and Clerc, Maureen and Congedo, Marco and Rakotomamonjy, Alain and Yger, Florian},
  title   = {A Review of Classification Algorithms for {EEG}-Based Brain--Computer Interfaces: A 10 Year Update},
  journal = {Journal of Neural Engineering},
  year    = {2018},
  volume  = {15},
  number  = {3},
  pages   = {031005},
  doi     = {10.1088/1741-2552/aab2f2}
}

@article{schirrmeister2017deep,
  author  = {Schirrmeister, Robin Tibor and Springenberg, Jost Tobias and Fiederer, Lukas Dominik Josef and Glasstetter, Martin and Eggensperger, Katharina and Tangermann, Michael and Hutter, Frank and Burgard, Wolfram and Ball, Tonio},
  title   = {Deep Learning with Convolutional Neural Networks for {EEG} Decoding and Visualization},
  journal = {Human Brain Mapping},
  year    = {2017},
  volume  = {38},
  number  = {11},
  pages   = {5391--5420},
  doi     = {10.1002/hbm.23730}
}

@article{roy2019systematic,
  author  = {Roy, Yannick and Banville, Hubert and Albuquerque, Isabela and Gramfort, Alexandre and Falk, Tiago H. and Faubert, Jocelyn},
  title   = {Deep Learning-Based Electroencephalography Analysis: A Systematic Review},
  journal = {Journal of Neural Engineering},
  year    = {2019},
  volume  = {16},
  number  = {5},
  pages   = {051001},
  doi     = {10.1088/1741-2552/ab260c}
}

@article{haufe2014interpretation,
  author  = {Haufe, Stefan and Meinecke, Frank and G{\"o}rgen, Kai and D{\"a}hne, Sven and Haynes, John-Dylan and Blankertz, Benjamin and Bie{\ss}mann, Felix},
  title   = {On the Interpretation of Weight Vectors of Linear Models in Multivariate Neuroimaging},
  journal = {NeuroImage},
  year    = {2014},
  volume  = {87},
  pages   = {96--110},
  doi     = {10.1016/j.neuroimage.2013.10.067}
}

@article{kaplan2020scaling,
  author  = {Kaplan, Jared and McCandlish, Sam and Henighan, Tom and Brown, Tom B. and Chess, Benjamin and Child, Rewon and Gray, Scott and Radford, Alec and Wu, Jeffrey and Amodei, Dario},
  title   = {Scaling Laws for Neural Language Models},
  journal = {arXiv preprint arXiv:2001.08361},
  year    = {2020},
  eprint  = {2001.08361},
  archivePrefix = {arXiv},
  primaryClass  = {cs.LG}
}

@inproceedings{frankle2019lottery,
  author    = {Frankle, Jonathan and Carbin, Michael},
  title     = {The Lottery Ticket Hypothesis: Finding Sparse, Trainable Neural Networks},
  booktitle = {International Conference on Learning Representations},
  year      = {2019}
}

@article{song2023eegconformer,
  author  = {Song, Yonghao and Zheng, Qingqing and Liu, Bingchuan and Gao, Xiaorong},
  title   = {{EEG} Conformer: Convolutional Transformer for {EEG} Decoding and Visualization},
  journal = {IEEE Transactions on Neural Systems and Rehabilitation Engineering},
  year    = {2023},
  volume  = {31},
  pages   = {710--719},
  doi     = {10.1109/TNSRE.2022.3230250}
}

@inproceedings{charbonnier1994,
  author    = {Charbonnier, Pierre and Blanc-F{\'e}raud, Laure and Aubert, Gilles and Barlaud, Michel},
  title     = {Two Deterministic Half-Quadratic Regularization Algorithms for Computed Imaging},
  booktitle = {Proceedings of the IEEE International Conference on Image Processing},
  year      = {1994},
  volume    = {2},
  pages     = {168--172}
}

@misc{xiang2025taskoriented,
  author        = {Xiang, Tian-Yu and Lei, Zheng and Zhou, Xiao-Hu and Xie, Xiao-Liang and Liu, Shi-Qi and Gui, Mei-Jiang and Ou, Hong-Yun and Huang, Xin-Zheng and Fu, Xin-Yi and Hou, Zeng-Guang},
  title         = {Task-Oriented Learning for Automatic {EEG} Denoising},
  year          = {2025},
  eprint        = {2509.14665},
  archivePrefix = {arXiv},
  primaryClass  = {eess.SP}
}

\end{document}


\articletype{Supplementary material}

\title{Supplementary Material: How Much Capacity Does EEG Denoising Need? Ultra-Compact Networks reveal Benchmark Saturation and Metric-Utility Gap}

\author{Jasmeet Singh Bindra$^{1}$\orcid{0009-0006-3105-0758}, Siddharth Panwar$^2$\orcid{0000-0002-9432-350X} and Shubhajit Roy Chowdhury$^{2,*}$\orcid{0000-0003-1878-6657}}

\affil{$^1$Indian Knowledge Systems and Mental Health Applications (IKSMHA) Center, Indian Institute of Technology Mandi, Himachal Pradesh, India}

\affil{$^2$School of Computing and Electrical Engineering, Indian Institute of Technology Mandi, Himachal Pradesh, India}

\affil{$^*$Author to whom any correspondence should be addressed.}

\email{D20459@students.iitmandi.ac.in; siddharth.panwar@iitmandi.ac.in; src@iitmandi.ac.in}

\begin{abstract}
This document provides supplementary figures, statistical summaries, and methodological details for the EEG denoising capacity analysis. It includes per-seed reconstruction plots, real-artifact qualitative examples, full relative-error curves, per-checkpoint downstream BCI diagnostics, removed-signal RMS summaries, fixed \(n=5\) EEGDenoiseNet width-sweep statistics, classical and tiny learned lower-anchor baselines, controlled EEGDN CNN and MicroWaveNet retraining results, Patch-Transformer capacity-sweep controls, expanded downstream classification details across five decoder families, calibration controls, convergence audits, and a CSP-feature task-aware loss pilot.
\end{abstract}

\section*{Supplementary Figures}

\begin{figure}[!htb]
    \centering
    \includegraphics[width=\textwidth]{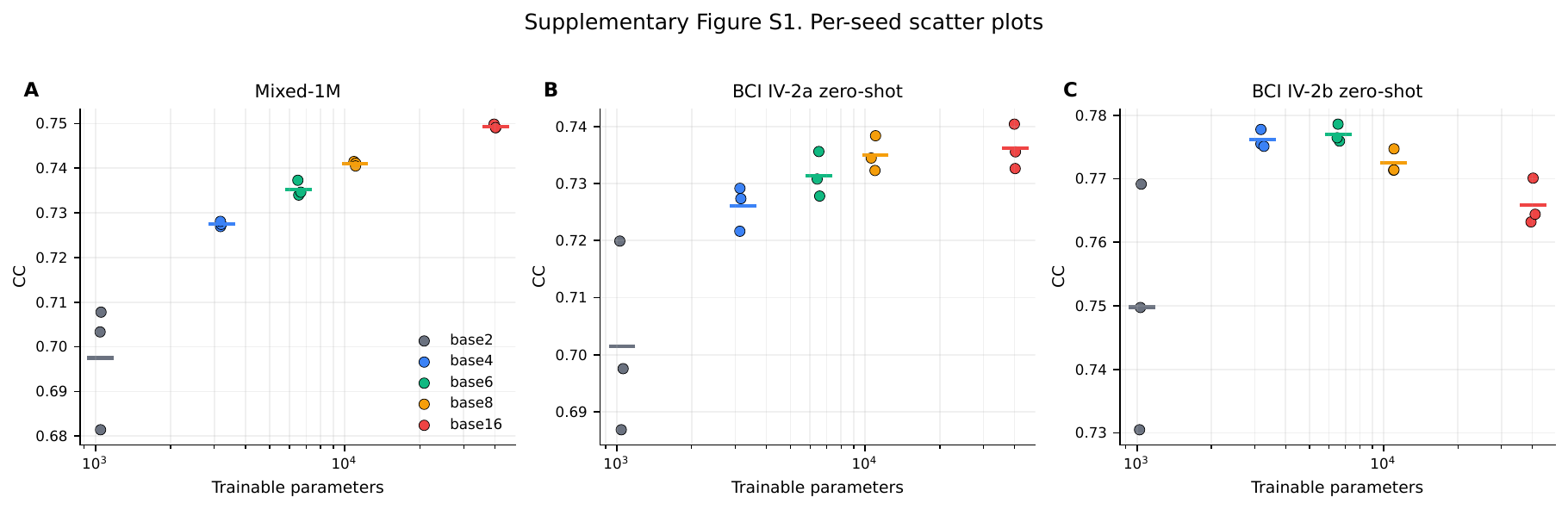}
    \caption{\textbf{Per-seed reconstruction results.}
    Per-seed scatter plots for the Mixed-1M, BCI Competition IV-2a, and BCI Competition IV-2b evaluations. Each point corresponds to one independently trained checkpoint. Width variants are shown by parameter count, and repeated points at the same width show variation across training seeds. These plots support the multi-seed reconstruction summaries reported in the main text.}
    \label{fig:supp_s1}
\end{figure}

\begin{figure}[!htb]
    \centering
    \includegraphics[width=\textwidth]{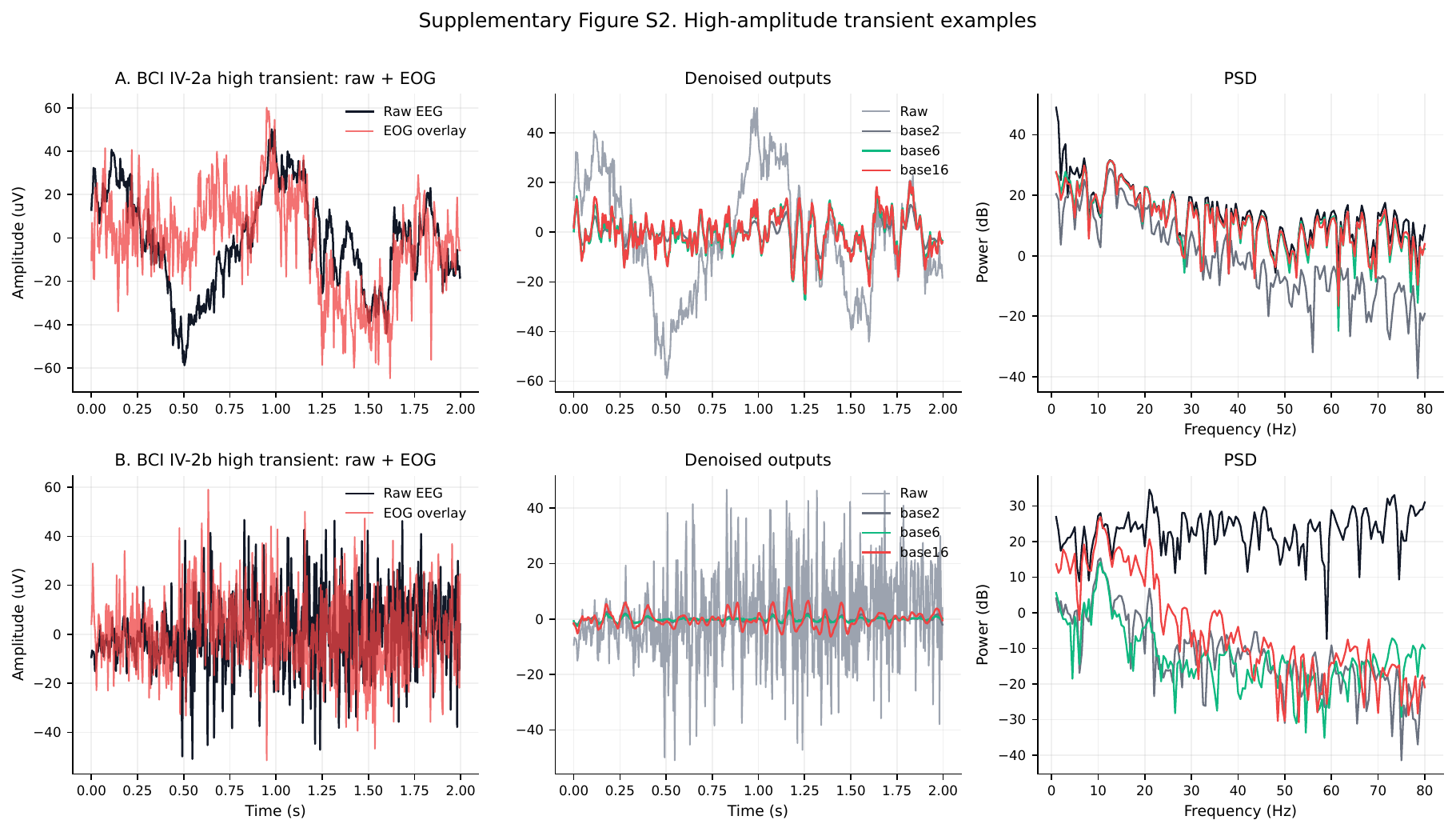}
    \caption{\textbf{Real-artifact high-amplitude transient examples.}
    Qualitative examples from high-amplitude transient windows in BCI Competition IV-2a and BCI Competition IV-2b. The windows were selected from continuous real BCI recordings without synthetic artifact injection. These examples complement the main-text qualitative figure by showing cases in which large EEG deflections are attenuated by the denoisers.}
    \label{fig:supp_s2}
\end{figure}

\begin{figure}[!htb]
    \centering
    \includegraphics[width=\textwidth]{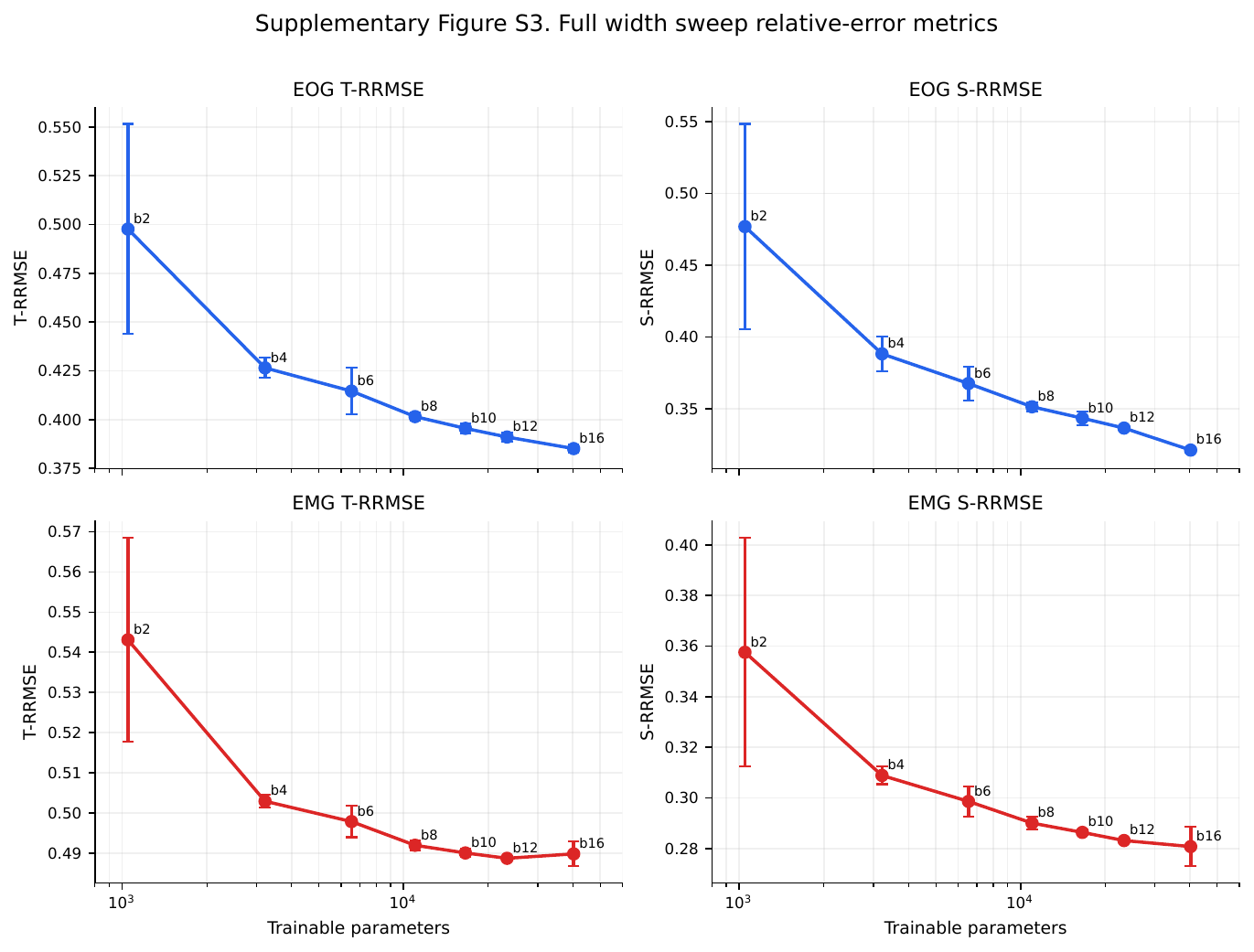}
    \caption{\textbf{Full relative-error width sweep on EEGDenoiseNet.}
    Temporal relative root mean square error (T-RRMSE) and spectral relative root mean square error (S-RRMSE) for EOG and EMG denoising on the EEGDenoiseNet benchmark. Width variants are plotted against trainable parameter count to show the behavior of relative-error metrics in addition to the correlation-coefficient results reported in the main text.}
    \label{fig:supp_s3}
\end{figure}

\begin{figure}[!htb]
    \centering
    \includegraphics[width=\textwidth]{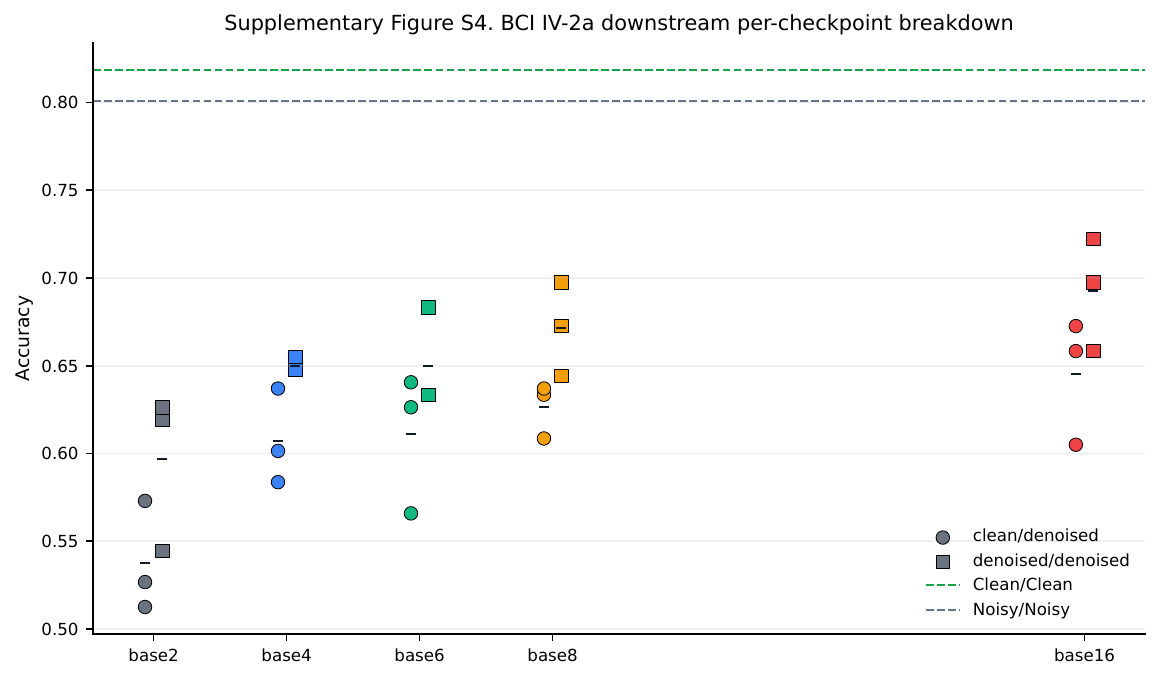}
    \caption{\textbf{Per-checkpoint downstream BCI IV-2a classification.}
    Per-checkpoint CSP+LDA motor-imagery classification diagnostics for clean/denoised and denoised/denoised conditions. Each point corresponds to one denoising checkpoint. The expanded all-subject, five-classifier downstream analysis is summarized in the main text and in Supplementary Tables S6--S10.}
    \label{fig:supp_s4}
\end{figure}

\begin{figure}[!htb]
    \centering
    \includegraphics[width=\textwidth]{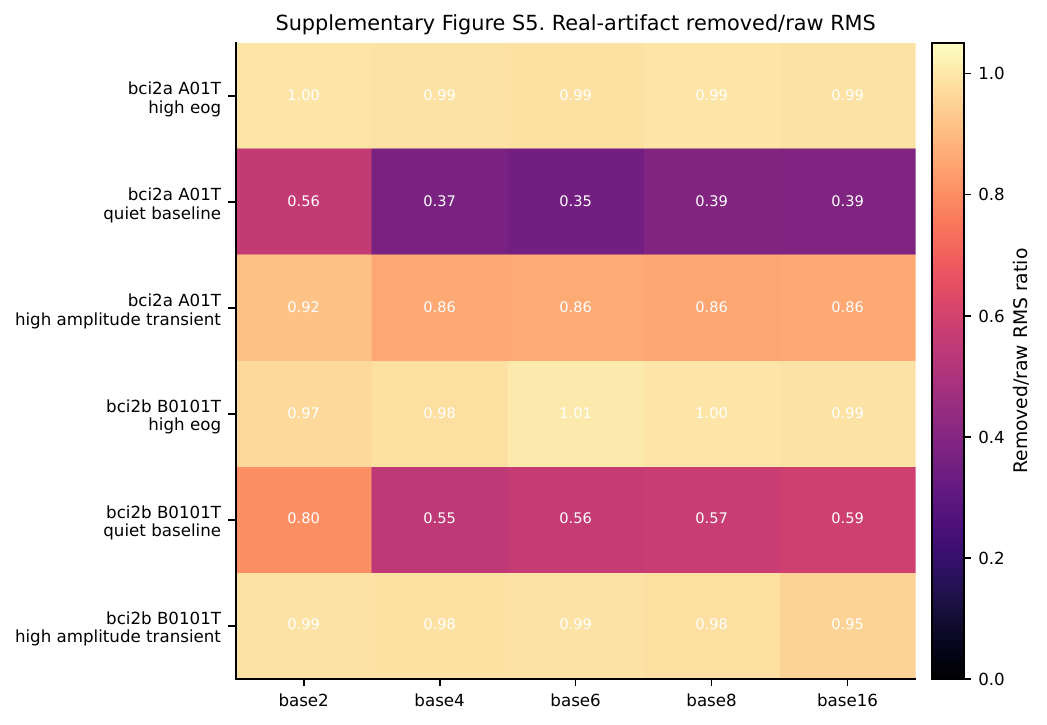}
    \caption{\textbf{Removed/raw RMS ratio for real-artifact windows.}
    Heatmap of the removed-signal RMS divided by raw-signal RMS across real-artifact qualitative conditions and controlled backbone width variants. Values near one indicate that the denoiser removed a signal component with RMS similar to the raw input, whereas lower values indicate weaker attenuation. These summaries are descriptive real-signal checks and are not clean-target denoising metrics.}
    \label{fig:supp_s5}
\end{figure}

\begin{table}[p]
\caption{\textbf{Fixed \(n=5\) EEGDenoiseNet width-sweep reconstruction statistics.}
Values are mean \(\pm\) standard deviation across five independent seeds for the key Pareto widths. Base10 and base12 were exploratory \(n=3\) widths in the main benchmark table and are not included here.}
\label{tab:supp_n5_reconstruction}
\centering
\footnotesize
\setlength{\tabcolsep}{4pt}
\begin{tabular}{lrrrrrr}
\hline
Dataset & Width & Params & CC & RMSE & T-RRMSE & S-RRMSE \\
\hline
EOG & base2 & 1.05K & \(0.871 \pm 0.014\) & \(0.243 \pm 0.026\) & \(0.496 \pm 0.038\) & \(0.474 \pm 0.051\) \\
EOG & base4 & 3.22K & \(0.898 \pm 0.002\) & \(0.201 \pm 0.002\) & \(0.426 \pm 0.004\) & \(0.386 \pm 0.009\) \\
EOG & base6 & 6.53K & \(0.904 \pm 0.003\) & \(0.195 \pm 0.007\) & \(0.412 \pm 0.009\) & \(0.365 \pm 0.009\) \\
EOG & base8 & 10.99K & \(0.907 \pm 0.001\) & \(0.189 \pm 0.001\) & \(0.402 \pm 0.003\) & \(0.353 \pm 0.003\) \\
EOG & base16 & 40.26K & \(0.915 \pm 0.001\) & \(0.179 \pm 0.001\) & \(0.385 \pm 0.002\) & \(0.320 \pm 0.005\) \\
\hline
EMG & base2 & 1.05K & \(0.822 \pm 0.007\) & \(0.266 \pm 0.015\) & \(0.539 \pm 0.019\) & \(0.353 \pm 0.033\) \\
EMG & base4 & 3.22K & \(0.837 \pm 0.001\) & \(0.243 \pm 0.001\) & \(0.503 \pm 0.001\) & \(0.307 \pm 0.004\) \\
EMG & base6 & 6.53K & \(0.840 \pm 0.001\) & \(0.240 \pm 0.002\) & \(0.497 \pm 0.003\) & \(0.299 \pm 0.005\) \\
EMG & base8 & 10.99K & \(0.843 \pm 0.001\) & \(0.237 \pm 0.001\) & \(0.492 \pm 0.001\) & \(0.291 \pm 0.003\) \\
EMG & base16 & 40.26K & \(0.843 \pm 0.002\) & \(0.236 \pm 0.001\) & \(0.491 \pm 0.003\) & \(0.285 \pm 0.009\) \\
\hline
\end{tabular}
\end{table}

\begin{table}[p]
\caption{\textbf{Controlled external-baseline retraining results.}
EEGDN CNN and MicroWaveNet were trained and evaluated under the same data split, loss, and metric pipeline used for the controlled backbone where possible. The MicroWaveNet controlled run is an architecture-only same-pipeline comparison and does not reproduce the original multi-loss Soft-DTW and wavelet-domain training recipe. Values are mean \(\pm\) standard deviation across three seeds.}
\label{tab:supp_controlled_baselines}
\centering
\footnotesize
\setlength{\tabcolsep}{4pt}
\begin{tabular}{llrrrrrr}
\hline
Model & Dataset & Params & CC & RMSE & T-RRMSE & S-RRMSE & SDR \\
\hline
EEGDN CNN & EOG & 8.46M & \(0.915 \pm 0.002\) & \(0.176 \pm 0.002\) & \(0.374 \pm 0.004\) & \(0.317 \pm 0.001\) & \(9.30 \pm 0.09\) \\
EEGDN CNN & EMG & 33.62M & \(0.789 \pm 0.002\) & \(0.283 \pm 0.001\) & \(0.606 \pm 0.004\) & \(0.376 \pm 0.006\) & \(5.33 \pm 0.04\) \\
MicroWaveNet & EOG & 53.50K & \(0.868 \pm 0.052\) & \(0.232 \pm 0.052\) & \(0.480 \pm 0.095\) & \(0.432 \pm 0.085\) & \(6.79 \pm 1.80\) \\
MicroWaveNet & EMG & 53.50K & \(0.799 \pm 0.003\) & \(0.276 \pm 0.002\) & \(0.561 \pm 0.003\) & \(0.365 \pm 0.002\) & \(5.60 \pm 0.07\) \\
\hline
\end{tabular}
\end{table}

\begin{table}[p]
\caption{\textbf{Classical and tiny learned EEGDenoiseNet lower-anchor baselines.}
Classical filters have zero trainable parameters. Learned FIR and tiny TCN rows are mean \(\pm\) standard deviation across seeds 42--44. The oracle artifact-regression row uses the stored synthetic artifact component and is included only as an oracle sanity check, not as a deployable baseline.}
\label{tab:supp_classical_tiny_baselines}
\centering
\footnotesize
\setlength{\tabcolsep}{3pt}
\resizebox{\textwidth}{!}{%
\begin{tabular}{llrrrrrr}
\hline
Task & Method & Params & CC & RMSE & T-RRMSE & S-RRMSE & SDR \\
\hline
EOG & noisy passthrough & 0 & 0.507 & 0.830 & 2.193 & 13.569 & -5.00 \\
EOG & bandpass 0.5--45 Hz & 0 & 0.492 & 0.830 & 2.186 & 13.567 & -5.02 \\
EOG & Haar wavelet soft-threshold & 0 & 0.388 & 0.866 & 2.241 & 13.343 & -5.45 \\
EOG & EOG high-pass 1--45 Hz & 0 & 0.552 & 0.695 & 1.826 & 11.512 & -3.26 \\
EOG & oracle artifact regression & 1 & 1.000 & 0.000 & 0.000 & 0.000 & 137.03 \\
EOG & learned FIR & 34 & \(0.784 \pm 0.001\) & \(0.298 \pm 0.000\) & \(0.638 \pm 0.000\) & \(0.703 \pm 0.002\) & \(4.25 \pm 0.01\) \\
EOG & tiny TCN & 365 & \(0.849 \pm 0.005\) & \(0.257 \pm 0.009\) & \(0.532 \pm 0.012\) & \(0.586 \pm 0.011\) & \(5.72 \pm 0.23\) \\
\hline
EMG & noisy passthrough & 0 & 0.572 & 0.754 & 1.903 & 2.242 & -3.00 \\
EMG & bandpass 0.5--45 Hz & 0 & 0.735 & 0.402 & 0.983 & 1.304 & 2.36 \\
EMG & Haar wavelet soft-threshold & 0 & 0.646 & 0.475 & 1.095 & 1.204 & 0.91 \\
EMG & EMG low-pass 40 Hz & 0 & 0.743 & 0.389 & 0.945 & 1.250 & 2.62 \\
EMG & oracle artifact regression & 1 & 1.000 & 0.000 & 0.000 & 0.000 & 139.11 \\
EMG & learned FIR & 34 & \(0.779 \pm 0.000\) & \(0.319 \pm 0.001\) & \(0.685 \pm 0.001\) & \(0.600 \pm 0.002\) & \(4.24 \pm 0.01\) \\
EMG & tiny TCN & 365 & \(0.797 \pm 0.003\) & \(0.283 \pm 0.002\) & \(0.595 \pm 0.004\) & \(0.413 \pm 0.004\) & \(5.35 \pm 0.07\) \\
\hline
\end{tabular}
}
\end{table}

\begin{table}[p]
\caption{\textbf{Patch-Transformer EEGDenoiseNet capacity sweep.}
Patch-Transformer denoisers were trained under the same controlled clean-reconstruction pipeline as the controlled backbone. Values are mean \(\pm\) standard deviation across seeds 42--44.}
\label{tab:supp_patch_transformer}
\centering
\footnotesize
\setlength{\tabcolsep}{3pt}
\resizebox{\textwidth}{!}{%
\begin{tabular}{llrrrrrr}
\hline
Task & Variant & Params & CC & RMSE & T-RRMSE & S-RRMSE & SDR \\
\hline
EOG & d8 & 1.35K & \(0.827 \pm 0.002\) & \(0.269 \pm 0.002\) & \(0.558 \pm 0.004\) & \(0.570 \pm 0.005\) & \(5.26 \pm 0.07\) \\
EOG & d16 & 4.87K & \(0.869 \pm 0.002\) & \(0.231 \pm 0.002\) & \(0.486 \pm 0.005\) & \(0.456 \pm 0.005\) & \(6.63 \pm 0.09\) \\
EOG & d24 & 10.56K & \(0.880 \pm 0.002\) & \(0.220 \pm 0.002\) & \(0.463 \pm 0.005\) & \(0.438 \pm 0.005\) & \(7.09 \pm 0.09\) \\
EOG & d32 & 18.43K & \(0.884 \pm 0.002\) & \(0.216 \pm 0.002\) & \(0.455 \pm 0.003\) & \(0.426 \pm 0.002\) & \(7.24 \pm 0.06\) \\
EOG & d48 & 40.71K & \(0.891 \pm 0.001\) & \(0.210 \pm 0.001\) & \(0.442 \pm 0.003\) & \(0.413 \pm 0.004\) & \(7.51 \pm 0.06\) \\
\hline
EMG & d8 & 1.35K & \(0.776 \pm 0.003\) & \(0.281 \pm 0.001\) & \(0.600 \pm 0.003\) & \(0.501 \pm 0.006\) & \(5.42 \pm 0.05\) \\
EMG & d16 & 4.87K & \(0.812 \pm 0.003\) & \(0.260 \pm 0.002\) & \(0.550 \pm 0.005\) & \(0.395 \pm 0.013\) & \(6.12 \pm 0.07\) \\
EMG & d24 & 10.56K & \(0.824 \pm 0.002\) & \(0.252 \pm 0.001\) & \(0.530 \pm 0.003\) & \(0.347 \pm 0.005\) & \(6.40 \pm 0.05\) \\
EMG & d32 & 18.43K & \(0.828 \pm 0.002\) & \(0.248 \pm 0.001\) & \(0.521 \pm 0.004\) & \(0.333 \pm 0.004\) & \(6.53 \pm 0.05\) \\
EMG & d48 & 40.71K & \(0.831 \pm 0.001\) & \(0.246 \pm 0.000\) & \(0.515 \pm 0.001\) & \(0.325 \pm 0.003\) & \(6.63 \pm 0.02\) \\
\hline
\end{tabular}
}
\end{table}

\begin{table}[p]
\caption{\textbf{Segmented regression knee analysis for EEGDenoiseNet CC.}
A descriptive segmented regression was fit to CC versus \(\log_{10}\) parameter count for the fixed \(n=5\) Pareto widths. Bootstrap knee counts report the selected elbow base over 5000 bootstrap resamples.}
\label{tab:supp_saturation_knee}
\centering
\footnotesize
\begin{tabular}{lrrrr}
\hline
Dataset & Elbow base & Params & Post-elbow slope & 95\% bootstrap CI \\
\hline
EOG & base4 & 3.22K & 0.0151 & 0.0123--0.0163 \\
EMG & base4 & 3.22K & 0.0044 & \(-0.0003\)--0.0056 \\
\hline
\end{tabular}
\par\smallskip
\raggedright\scriptsize Bootstrap knee counts were EOG: base4 4842, base6 153, base8 5; EMG: base4 4169, base6 821, base8 10.
\end{table}

\begin{table}[p]
\caption{\textbf{Downstream BCI IV-2a classifier baselines.}
Values are mean \(\pm\) standard deviation across all nine BCI Competition IV-2a subjects. The noisy/noisy condition is the practical no-denoising baseline for the synthetic EOG-contamination protocol.}
\label{tab:supp_downstream_baselines}
\centering
\footnotesize
\begin{tabular}{lccc}
\hline
Classifier & Clean/Clean & Clean/Noisy & Noisy/Noisy \\
\hline
CSP+LDA & \(0.607 \pm 0.176\) & \(0.570 \pm 0.186\) & \(0.612 \pm 0.173\) \\
ShallowFBCSPNet & \(0.537 \pm 0.150\) & \(0.436 \pm 0.123\) & \(0.497 \pm 0.147\) \\
EEGConformer & \(0.538 \pm 0.163\) & \(0.426 \pm 0.117\) & \(0.484 \pm 0.173\) \\
EEGNet & \(0.496 \pm 0.165\) & \(0.445 \pm 0.155\) & \(0.423 \pm 0.131\) \\
Deep4Net & \(0.451 \pm 0.161\) & \(0.426 \pm 0.165\) & \(0.391 \pm 0.123\) \\
\hline
\end{tabular}
\end{table}

\begin{table}[p]
\caption{\textbf{Complete matched denoised/denoised downstream BCI IV-2a matrix.}
Each cell reports accuracy as mean \(\pm\) standard deviation across nine subjects, followed by uncorrected one-sided Wilcoxon \(p\) and BH-FDR \(q\) across all 25 width-by-classifier comparisons.}
\label{tab:supp_downstream_matrix}
\centering
\tiny
\setlength{\tabcolsep}{3pt}
\resizebox{\textwidth}{!}{%
\begin{tabular}{lccccc}
\hline
Width & CSP+LDA & ShallowFBCSPNet & EEGConformer & EEGNet & Deep4Net \\
\hline
base2 & \(0.491 \pm 0.130\) (0.0059/0.0488) & \(0.454 \pm 0.151\) (0.0273/0.0854) & \(0.424 \pm 0.125\) (0.0195/0.0814) & \(0.413 \pm 0.118\) (0.3672/0.4831) & \(0.395 \pm 0.122\) (0.6328/0.7191) \\
base4 & \(0.522 \pm 0.147\) (0.0059/0.0488) & \(0.473 \pm 0.156\) (0.0488/0.1221) & \(0.450 \pm 0.146\) (0.0820/0.1578) & \(0.414 \pm 0.129\) (0.2852/0.3961) & \(0.403 \pm 0.126\) (0.7871/0.8199) \\
base6 & \(0.536 \pm 0.159\) (0.0098/0.0488) & \(0.476 \pm 0.150\) (0.0645/0.1343) & \(0.450 \pm 0.151\) (0.0645/0.1343) & \(0.417 \pm 0.129\) (0.2480/0.3876) & \(0.395 \pm 0.117\) (0.4551/0.5418) \\
base8 & \(0.529 \pm 0.174\) (0.0098/0.0488) & \(0.475 \pm 0.153\) (0.0488/0.1221) & \(0.469 \pm 0.159\) (0.1797/0.2995) & \(0.415 \pm 0.138\) (0.4102/0.5127) & \(0.403 \pm 0.121\) (0.7520/0.8173) \\
base16 & \(0.547 \pm 0.174\) (0.0098/0.0488) & \(0.477 \pm 0.155\) (0.0273/0.0854) & \(0.466 \pm 0.154\) (0.1504/0.2686) & \(0.411 \pm 0.136\) (0.2852/0.3961) & \(0.409 \pm 0.132\) (0.8496/0.8496) \\
\hline
\end{tabular}
}
\end{table}

\begin{table}[p]
\caption{\textbf{CSP+LDA sensitivity to post-denoising normalization and covariance calibration.}
Each row reports the best fixed-width matched denoised/denoised setting for one calibration transform. Values are mean \(\pm\) standard deviation across nine subjects. Wilcoxon \(p\)-values compare denoised/denoised against noisy/noisy with the one-sided lower alternative.}
\label{tab:supp_calibration_sensitivity}
\centering
\footnotesize
\resizebox{\textwidth}{!}{%
\begin{tabular}{lcccccc}
\hline
Calibration & Best width & Accuracy & \(\Delta\) vs noisy/noisy & Subjects below & \(p\) & BH \(q\) \\
\hline
None & base16 & \(0.547 \pm 0.174\) & \(-0.065 \pm 0.060\) & 7/9 & 0.0098 & 0.0122 \\
Train-channel z-score & base16 & \(0.548 \pm 0.172\) & \(-0.064 \pm 0.059\) & 7/9 & 0.0098 & 0.0122 \\
Raw-variance rescale & base16 & \(0.546 \pm 0.173\) & \(-0.066 \pm 0.059\) & 7/9 & 0.0098 & 0.0122 \\
Covariance recolor & base8 & \(0.540 \pm 0.171\) & \(-0.072 \pm 0.061\) & 7/9 & 0.0098 & 0.0122 \\
Trial z-score & base6 & \(0.570 \pm 0.167\) & \(-0.042 \pm 0.064\) & 6/9 & 0.0645 & 0.0645 \\
\hline
\end{tabular}
}
\end{table}

\begin{table}[p]
\caption{\textbf{Long-budget Deep4Net and EEGConformer convergence audit.}
The audit used 200 training epochs for the two neural decoders most affected by convergence concerns. Values are mean \(\pm\) standard deviation across nine subjects.}
\label{tab:supp_neural_convergence}
\centering
\footnotesize
\resizebox{\textwidth}{!}{%
\begin{tabular}{lccccc}
\hline
Classifier & Noisy/noisy & Denoised width & Denoised/denoised & \(\Delta\) vs noisy/noisy & \(p\) / BH \(q\) \\
\hline
Deep4Net & \(0.430 \pm 0.162\) & base16 & \(0.440 \pm 0.156\) & \(+0.010 \pm 0.035\) & 0.8203 / 0.8203 \\
EEGConformer & \(0.515 \pm 0.229\) & base8 & \(0.479 \pm 0.190\) & \(-0.036 \pm 0.071\) & 0.1504 / 0.3008 \\
\hline
\end{tabular}
}
\end{table}

\begin{table}[p]
\caption{\textbf{Task-aware CSP-feature preservation pilot.}
A frozen CSP-feature auxiliary loss was added to the 22-channel multichannel denoiser objective. Stronger weighting worsened matched CSP+LDA performance relative to reconstruction-only denoising. Values are mean \(\pm\) standard deviation across nine BCI IV-2a subjects.}
\label{tab:supp_csp_feature_pilot}
\centering
\footnotesize
\resizebox{\textwidth}{!}{%
\begin{tabular}{lccccc}
\hline
Objective & CSP \(\lambda\) & Params & Accuracy & \(\Delta\) vs noisy/noisy & \(p\) / BH \(q\) \\
\hline
Reconstruction only & 0 & 602.99K & \(0.570 \pm 0.176\) & \(-0.043 \pm 0.069\) & 0.0645 / 0.0645 \\
Reconstruction + CSP features & 0.1 & 602.99K & \(0.551 \pm 0.180\) & \(-0.061 \pm 0.067\) & 0.0273 / 0.0410 \\
Reconstruction + CSP features & 1.0 & 602.99K & \(0.510 \pm 0.178\) & \(-0.102 \pm 0.082\) & 0.0098 / 0.0293 \\
\hline
\end{tabular}
}
\end{table}

\begin{table}[p]
\caption{\textbf{Architecture design-justification ablation.} Values are mean \(\pm\) standard deviation over three seeds.}
\label{tab:supp_ablation}
\centering
\footnotesize
\begin{tabular}{lcccc}
\hline
Variant & Params & EOG CC & EMG CC & Notes \\
\hline
Backbone only & 40.24K & \(0.915 \pm 0.001\) & \(0.838 \pm 0.002\) & DSConv backbone \\
+ Artifact loss & 40.25K & \(0.913 \pm 0.002\) & \(0.838\) & Clean + artifact target \\
+ Attention & 40.26K & \(0.915 \pm 0.001\) & \(0.843 \pm 0.002\) & ECA attention \\
+ Spectral loss & 40.24K & \(0.908 \pm 0.001\) & \(0.838 \pm 0.001\) & Spectral objective \\
\hline
\end{tabular}
\end{table}

\begin{figure}[!htb]
    \centering
    \begin{minipage}[t]{0.48\textwidth}
        \centering
        \includegraphics[width=\textwidth]{fig02_ablation_summary}
        \vspace{-2mm}
        {\footnotesize (a) Design-justification ablation}
    \end{minipage}
    \hfill
    \begin{minipage}[t]{0.48\textwidth}
        \centering
        \includegraphics[width=\textwidth]{fig04_snr_stratified}
        \vspace{-2mm}
        {\footnotesize (b) SNR-stratified Mixed-1M CC}
    \end{minipage}

    \vspace{4mm}

    \begin{minipage}[t]{0.65\textwidth}
        \centering
        \includegraphics[width=\textwidth]{fig05_artifact_recipe_heatmap}
        \vspace{-2mm}
        {\footnotesize (c) Artifact recipe heatmap}
    \end{minipage}
    \caption{\textbf{Additional reconstruction diagnostics.}
    (a) Ablation summary comparing backbone-only, attention, spectral-loss, and artifact-loss variants for EOG and EMG reconstruction.
    (b) SNR-stratified Mixed-1M CC across width variants; error bars denote variation across training seeds.
    (c) Artifact recipe heatmap for Mixed-1M; rows correspond to artifact recipes and columns to width variants; color denotes CC.}
    \label{fig:supp_s6}
\end{figure}

\clearpage
\FloatBarrier
\section*{Additional Methodological Details}

\subsection*{Mixed-Artifact Recipe Generation (Extended Detail)}

This section provides extended procedural detail for the mixed-artifact corpus described in Section 3.3.2 of the main paper. The large-scale mixed-artifact corpus was generated from EEGDenoiseNet-derived source pools at 256 Hz with two-second segments of length \(T=512\). The source pools contained 4,514 clean EEG segments of 512 samples, 3,400 EOG segments of 512 samples, and 5,598 EMG segments of 1,024 samples. EMG segments were downsampled from 512 Hz to 256 Hz before mixing. Clean EEG, EOG, and EMG source indices were split independently into train, validation, and test pools before any mixture generation. This source-level disjoint split prevents the same clean EEG, EOG, or EMG source segment from appearing across train, validation, and test mixtures.

Each mixed-artifact segment was generated by sampling a clean EEG segment \(x\) and then sampling one of seven artifact recipes with probabilities:
\[
\begin{array}{ll}
\mathrm{EOG}: 0.25, & \mathrm{EMG}: 0.25,\\
\mathrm{EOG+EMG}: 0.20, & \mathrm{EMG+LINE}: 0.10,\\
\mathrm{EOG+LINE}: 0.10, & \mathrm{EOG+EMG+LINE}: 0.05,\\
\mathrm{EOG+EMG+LINE+ECG}: 0.05.
\end{array}
\]
Recipe identifiers were stored as metadata: 0 for EOG, 1 for EMG, 2 for EOG+EMG, 3 for EMG+LINE, 4 for EOG+LINE, 5 for EOG+EMG+LINE, and 6 for EOG+EMG+LINE+ECG.

For EOG-containing recipes, one EOG segment was sampled from the split-specific EOG pool, randomly polarity-flipped with probability 0.5, and pre-scaled by a factor drawn uniformly from \([0.7,1.3]\). For EMG-containing recipes, one EMG segment was sampled from the split-specific EMG pool and pre-scaled by a factor drawn uniformly from \([0.6,1.5]\). Line-noise recipes added a normalized sinusoidal artifact with a 50 Hz fundamental with probability 0.85 and a 60 Hz fundamental otherwise. Harmonics and slow amplitude modulation were included. ECG recipes added a normalized synthetic pulse train with heart rate sampled from 50--110 beats per minute, Gaussian QRS pulses, and optional T-wave components. Electrode drift and pop noise were added independently with probability 0.35 to any recipe using smoothed random-walk drift, sparse exponentially decaying impulses, and white sensor noise.

The unscaled artifact sum \(a_0\) was scaled to a sampled signal-to-noise ratio (SNR). The SNR distribution was a two-part mixture: with probability 0.70, SNR was sampled uniformly from \([-7,2]\) dB, and with probability 0.30 from \([-12,-7]\) dB. The scaled artifact was
\[
a = \lambda a_0, \qquad
\lambda = \frac{\mathrm{RMS}(x)}{10^{\mathrm{SNR}/20}\,\mathrm{RMS}(a_0)} ,
\]
with a small numerical floor on the denominator. The noisy segment was \(y=x+a\). The noisy signal, clean target, and artifact target were normalized by \(\sigma_y=\mathrm{std}(y)\):
\[
\hat{y}=\frac{y}{\sigma_y}, \qquad
\hat{x}=\frac{x}{\sigma_y}, \qquad
\hat{a}=\frac{a}{\sigma_y}.
\]
The final corpus contained 800,000 training, 100,000 validation, and 100,000 test segments, stored in chunks of 10,000 samples with metadata fields for SNR, recipe identifier, source indices, auxiliary artifact flags, and the SNR scaling factor.

\subsection*{BCI Zero-Shot Denoising Protocol}

The BCI zero-shot reconstruction evaluation used BCI-derived semi-synthetic mixtures and did not fine-tune any denoising model. For BCI Competition IV-2a, the source file was A01T.mat. The clean EEG source was channel index 9, and the EOG artifact source was channel index 22. For BCI Competition IV-2b, the source file was B0101T.gdf. The clean EEG source was EEG:C3, with fallback index 0 if the channel name was unavailable, and the EOG artifact source was EOG:ch01, selected as the first channel whose name contained the substring EOG.

Continuous signals were resampled from 250 Hz to 256 Hz and segmented into two-second windows of 512 samples. Candidate windows were scored using EOG-band energy after 0.5--10 Hz bandpass filtering. Clean EEG windows were selected from the lowest 20\% of the EOG-energy distribution, and artifact windows from the highest 10\%. For each dataset, 5000 semi-synthetic test samples were generated with random seed 42. Clean and artifact windows were sampled independently from the corresponding pools, mixed at SNR values sampled uniformly from \([-6,2]\) dB, and normalized by \(\sigma_y=\mathrm{std}(y)\). Metrics were computed in the normalized domain against the known clean BCI EEG segment. This protocol evaluates zero-shot denoising transfer on BCI-derived signals; it is not a downstream classification protocol.

\subsection*{Downstream Classification Protocol}

Downstream utility was evaluated on BCI Competition IV-2a using all nine subjects (A01--A09) and the official training/evaluation session separation. For each subject, A0xT was used for classifier training and A0xE for held-out testing. Motor-imagery epochs were extracted from 2.5--6.0 s after trial onset and resampled from 250 Hz to 256 Hz. The first 22 EEG channels were used for classification, and channel index 22 served as the EOG source for synthetic contamination. Noisy training and test epochs were generated by adding the trial-matched EOG waveform to each EEG channel at an SNR sampled uniformly from \([-6,2]\) dB. The test noisy set used seed 42, and the train noisy set used seed 1042.

The CSP+LDA pipeline used 8--30 Hz bandpass filtering, 8 CSP components, OAS covariance regularization, log-transformed CSP features, concatenated covariance estimation, and no trace normalization. LDA used the \texttt{lsqr} solver with automatic shrinkage. Neural decoders used official Braindecode implementations of EEGNet, ShallowFBCSPNet, Deep4Net, and EEGConformer. All neural decoders were trained for 80 epochs using AdamW with learning rate \(10^{-3}\), weight decay \(10^{-4}\), and batch size 64. The condition matrix consisted of clean/clean, clean/noisy, noisy/noisy, clean/denoised, and denoised/denoised evaluations. In clean/denoised rows, classifiers were trained on clean A0xT and tested on denoised A0xE. In denoised/denoised rows, classifiers were trained and tested on denoised noisy epochs produced by the same denoising checkpoint.

For the real non-synthetic CSP+LDA experiment, denoisers were applied directly to recorded BCI IV-2a trials without synthetic EOG injection. The comparison was raw/raw versus matched denoised/denoised training and testing across all nine subjects. This analysis was added to test whether the spatial-filter utility gap persisted when denoising operated on naturally recorded EEG rather than on synthetically contaminated trials.

For contamination-type robustness, the same CSP+LDA protocol was repeated with EOG, EMG, and EOG+EMG+LINE synthetic contaminants. EOG templates were taken from the BCI IV-2a EOG channel. EMG templates were sampled from the EEGDenoiseNet raw EMG artifact pool and downsampled to 256 Hz. Mixed EOG+EMG+LINE contamination combined BCI EOG, EEGDenoiseNet EMG, and synthetic 50/60 Hz line noise before scaling the final artifact mixture to the target SNR. All recipes used the same subject set, trial window, artifact-flag exclusion rule, and CSP+LDA configuration as the main downstream experiment.

For the post-denoising calibration sensitivity analysis, CSP+LDA was evaluated after five transformations of the denoised signals: no calibration, train-channel z-scoring, raw-variance rescaling, covariance recoloring, and per-trial z-scoring. The comparison was always matched denoised/denoised versus noisy/noisy, using the same all-subject synthetic EOG protocol.

For the multichannel denoiser control, subject-specific 22-channel DSConv U-Net denoisers were trained on A0xT and evaluated on A0xE for base6 and base16. The multichannel control used the same reconstruction objective and AdamW settings as the main denoisers, with 120 training epochs, batch size 64, and seeds 42--44. This analysis was designed only to test whether covariance-aware denoising mitigated the CSP+LDA gap; it was not used as a replacement for the single-channel reconstruction benchmark.

For the neural-decoder convergence audit, Deep4Net and EEGConformer were retrained for 200 epochs under the same AdamW settings as the main 80-epoch neural-decoder protocol. The audit used the best fixed width from the original neural-decoder summary: base16 for Deep4Net and base8 for EEGConformer.

For the CSP-feature task-aware pilot, subject-specific 22-channel base16 DSConv U-Net denoisers were trained on A0xT and evaluated on A0xE. CSP filters were fit on the noisy/noisy training data and then frozen. The auxiliary term matched log-variance CSP features between denoised and clean targets after the same 8--30 Hz motor-imagery preprocessing used by the CSP+LDA pipeline. The total loss was the reconstruction loss plus \(\lambda\) times the CSP-feature loss, with \(\lambda \in \{0,0.1,1.0\}\). All pilot models used AdamW with learning rate \(10^{-3}\), weight decay \(10^{-4}\), batch size 64, 120 training epochs, and seeds 42--44.

\subsection*{Statistical Testing}

For downstream matched denoised/denoised comparisons, statistical significance was assessed with one-sided Wilcoxon signed-rank tests across subjects, with the alternative hypothesis that denoised accuracy is lower than the corresponding noisy/noisy or raw/raw baseline. The synthetic-contamination classifier-family analysis reports uncorrected \(p\)-values together with Bonferroni and/or Benjamini--Hochberg false-discovery-rate corrections. The real non-synthetic CSP+LDA experiment reports Benjamini--Hochberg false-discovery-rate \(q\)-values across the five width variants.

\subsection*{Compute Profiling}

Compute profiling was performed on 512-sample single-channel inputs for the controlled backbone width variants and EEGDenoiseNet CNN/RNN baselines. Batch sizes 1 and 256 were profiled on CPU and CUDA. CPU profiling used one thread. For each model, the profiler measured trainable parameter count, serialized PyTorch \texttt{state\_dict} size, multiply-accumulate operations (MACs), FLOPs, inference latency, throughput, and peak CUDA memory.

MACs were counted with forward hooks on \texttt{Conv1d}, \texttt{ConvTranspose1d}, \texttt{Linear}, and \texttt{LSTM} layers. FLOPs were reported as \(2\times\) MACs. Latency was measured with 30 warm-up iterations followed by 100 timed iterations. CUDA measurements called \texttt{torch.cuda.synchronize()} before and during timing to avoid asynchronous timing error. Peak GPU memory was measured by resetting CUDA peak memory statistics before profiling and then reading \texttt{torch.cuda.max\_memory\_allocated()} after the timed forward passes. For CPU rows, peak memory was reported as zero because CUDA allocator statistics are not defined.